%% file: main.tex
\documentclass[10pt,journal,compsoc]{IEEEtran}

\input{packages.tex}
\input{macros.tex}
\input{acronyms.tex}

\begin{document}
	
\title{Variational Hierarchical Mixtures for \\ Probabilistic Learning of Inverse Dynamics}

\author{Hany~Abdulsamad, Peter~Nickl, Pascal~Klink,
	and~Jan~Peters,~\IEEEmembership{Fellow,~IEEE.}%
	\IEEEcompsocitemizethanks{\IEEEcompsocthanksitem H.~Abdulsamad is with the Department of Electrical Engineering and Automation at Aalto University, Finland. P.~Nickl is with the RIKEN Center for Advanced Intelligence Project, Japan. P.~Klink, and J.~Peters are with the Department of Computer Science at the Technical University of Darmstadt, Germany.} %
}

\markboth{Journal of \LaTeX\ Class Files,~Vol.~14, No.~8, September~2021}%
{Shell \MakeLowercase{\emph{et al.}}: Bare Demo of IEEEtran.cls for Computer Society Journals}

\IEEEtitleabstractindextext{%
	\begin{abstract}
		Well-calibrated probabilistic regression models are a crucial learning component in robotics applications as datasets grow rapidly and tasks become more complex. Unfortunately, classical regression models are usually either probabilistic kernel machines with a flexible structure that does not scale gracefully with data or deterministic and vastly scalable automata, albeit with a restrictive parametric form and poor regularization. In this paper, we consider a probabilistic hierarchical modeling paradigm that combines the benefits of both worlds to deliver computationally efficient representations with inherent complexity regularization. The presented approaches are probabilistic interpretations of local regression techniques that approximate nonlinear functions through a set of local linear or polynomial units. Importantly, we rely on principles from Bayesian nonparametrics to formulate flexible models that adapt their complexity to the data and can potentially encompass an infinite number of components. We derive two efficient variational inference techniques to learn these representations and highlight the advantages of hierarchical infinite local regression models, such as dealing with non-smooth functions, mitigating catastrophic forgetting, and enabling parameter sharing and fast predictions. Finally, we validate this approach on large inverse dynamics datasets and test the learned models in real-world control scenarios.
	\end{abstract}
	
	\begin{IEEEkeywords}
		Hierarchical Local Regression, Generative Models, Dirichlet Process Mixtures, Inverse Dynamics Control
	\end{IEEEkeywords}
}

\maketitle

\IEEEdisplaynontitleabstractindextext
\IEEEpeerreviewmaketitle
\IEEEraisesectionheading{\section{Introduction}\label{sec:intro}}

\IEEEPARstart{P}{rincipled} data-driven, adaptive, and incremental learning is desirable in domains in which datasets are dynamic and accumulate slowly over time. For example, robots must build models of their dynamics and the environment as they interact with the world. Moreover, these models must be computationally efficient during both learning and evaluation. In the case of general-purpose robots, these models must incorporate different modalities of continuous and discrete stochastic random variables and possibly incorporate heteroscedastic noise \cite{todorov2005stochastic, buechler2019control}. Predominant and successful regression techniques, such as \gls{GPR} \cite{rasmussen2005gaussian}, \glspl{ANN} \cite{goodfellow2016deep}, and \gls{LR} \cite{wasserman2006all}, have a mixed set of properties that are useful in different scenarios. 

Gaussian process regression offers a principled Bayesian treatment that enables continual and incremental learning. Nonetheless, the $\emph{vanilla}$ formulation of \gls{GPR} \cite{rasmussen2005gaussian} suffered from many drawbacks that have been gradually addressed by recent research. Some of these drawbacks are the functional smoothness assumption \cite{calandra2016manifold, wilson2016deep, salimbeni2017doubly}, limitations when scaling to large datasets \cite{herbrich2003fast, titsias2009variational, cao2014generalized, deisenroth2015distributed, bauer2016understanding, matthews2017scalable} and difficulties modeling heteroscedasticity \cite{le2005heteroscedastic, kersting2007most, liu2020large}.

\begin{figure*}[t]
	\begin{minipage}[t]{0.45\textwidth}
		\centering
		\includegraphics{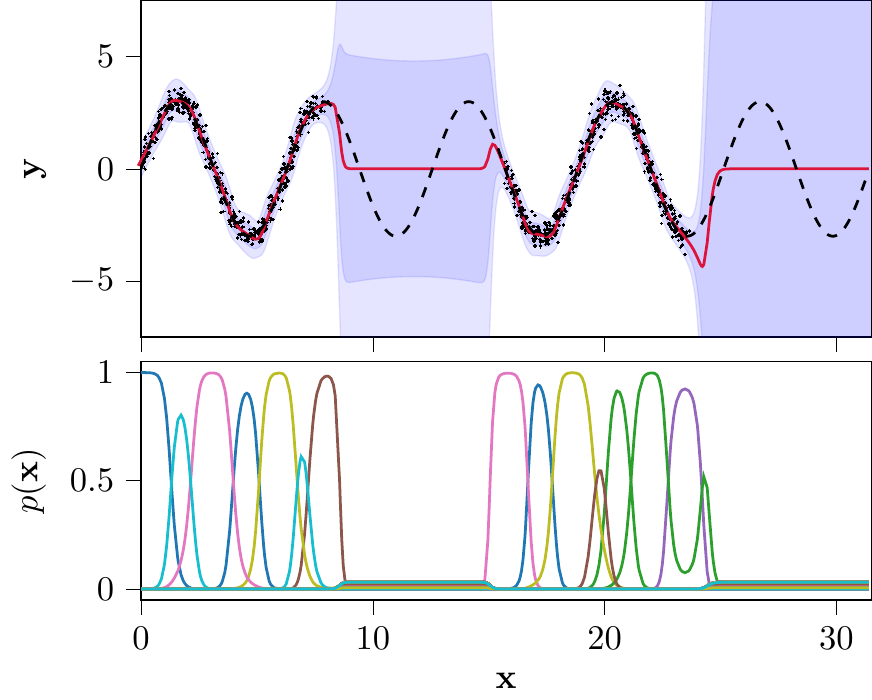}
	\end{minipage}\hfill
	\begin{minipage}[t]{0.45\textwidth}
		\centering		
		\includegraphics{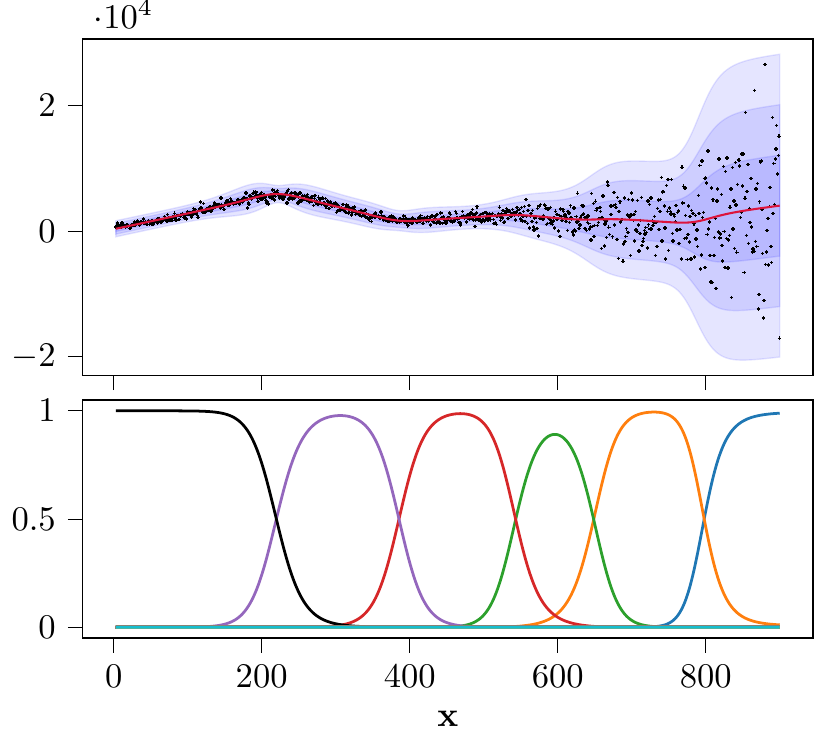}
	\end{minipage}	
	\vspace{-0.25cm}
	\caption{Left, gap data learned with \acf{ILR}. The top plot depicts the mean prediction (red) on the training data (dots) and the true mean function (dashed). The blue area represents the predictive uncertainty of two standard deviations. The predictive uncertainty in areas lacking training data is large, and the mean prediction falls back to the prior. Right, the \acs{CMB} dataset learned by \acs{ILR}. The top figure depicts the mean prediction (red) with a predictive uncertainty of three standard deviations (shaded blue). \acs{ILR} captures the heteroscedastic spread of the data with a handful of local regression models. The bottom plots show the activation of the models over the input space.}
	\label{fig:sine_cmb}
	\vspace{-0.35cm}
\end{figure*}

On the other hand, artificial neural networks have proven themselves as very powerful easy-to-train universal approximators. They are, however, still susceptible to over-parameterization \cite{frankle2018lottery} and catastrophic forgetting \cite{mccloskey1989catastrophic}. Moreover, despite major progress on the front of \glspl{BNN} \cite{neal1994bayesian, blundell2015weight, lakshminarayanan2017simple, khan2018fast, sun2019functional, watson2021latent, daxberger2021laplace}, new evidence suggests that issues regarding the accuracy of uncertainty quantification still need to be tackled \cite{wenzel2020good, foong2020expressiveness}.

Finally, local regression methods have had great success in the domain of robotics and control \cite{atkeson1997control, schaal1998constructive, schaal2002scalable, vijayakumar2005incremental}, because of their flexibility, ability to model hard nonlinearities, and to incorporate new data online. More generally, local regression is a family of generative \gls{MoE} techniques that take a basis-function approach to model the input density and automatically induce local model responsibilities \cite{moody1989fast, xu1994alternative, nelles1996basis, moerland1999classification}, see Figure~\ref{fig:sine_cmb}. In contrast, discriminative \glspl{MoE} rely on an explicitly input-conditioned gating to choose the local expert \cite{jacobs1991adaptive, jordan1994hierarchical, rasmussen2002infinite}. 

Two categories of \gls{LR} exist \cite{Ting2010Locally}, \emph{lazy} learners that maintain all seen data points in memory \cite{atkeson1997control, schaal2002scalable}, and \emph{memoryless} learners that compress data by constructing basis functions in the input space and fitting and storing locally parameterized regression models \cite{nelles1996basis}. Prominent examples of the latter include \gls{RFWR} \cite{schaal1998constructive} and \gls{LWPR} \cite{vijayakumar2005incremental}. However, these methods are often difficult to tune as they possess many hyperparameters. 

A limited attempt at a Bayesian treatment of \gls{LR} is made in \cite{ting2009bayesian} by constructing local nonparametric kernels and placing gamma priors on the kernel widths to alleviate the need to tune the basis functions. This approach leads to a localized GP formulation that needs to retain the training data in memory, again leading to the computational issues of vanilla \gls{GPR}. \Gls{LGR} is a further Bayesian generalization of \gls{LR} \cite{meier2014incremental}. The authors treat the local models in a Bayesian framework and couple them via the loss function that reinforces global coordination. Nonetheless, both approaches rely on heuristics and thresholds for adding and pruning local models and fall short of formulating a generative model over input and output.

Following this introduction, we believe that local regression with a proper generative Bayesian treatment can potentially serve as a powerful general-purpose function approximator. Moreover, as a probabilistic and efficient representation, it can drive many low-level applications in control and robotics that favor fast predictions and do not require deep representations. 

We introduce two probabilistic graphical mixture models for local regression in the upcoming sections. The first, \gls{ILR} \cite{abdulsamad2020variational}, is a generative formulation that relies on the paradigm of \gls{BNP} \cite{hjort2010bayesian} to automatically grow the mixture size based on observed data. This technique ultimately results in a general formulation of related methods that alleviates the need for any heuristic considerations. However, despite the effectiveness of \gls{ILR}, and like other local regression techniques that rely on locally linear or polynomial approximations, it maintains a one-to-one correspondence between the activations and local regression units. This effect limits the model's capacity to share parameters across the input space and often forces the generation of duplicate components, needlessly increasing the overall number of parameters. To address this limitation, we introduce \gls{HILR}, a multi-level development of \gls{ILR} that enables multi-modal activations of the same regression component, giving the model a structure that allows sharing of regression parameters across repeating local patterns in the data. This architecture increases the flexibility of the representation and contributes towards its compression.

For learning these models, we derive two general \gls{VB} schemes \cite{beal2003variational} that efficiently infer the posterior parameters and overcome the need for computationally heavy sampling methods. We benchmark the models on a range of tasks highlighting their strengths, such as dealing with heteroscedasticity, non-continuous functions, and multi-modal activation. Additionally, we test on large real-world high-dimensional datasets to benchmark learning the inverse dynamics of robotic manipulators. Most importantly, we deploy an instance of \gls{ILR} to perform inverse dynamics control on a real Barrett-WAM manipulator.

Alternative Bayesian extensions of the generative mixture of experts exist in the literature. A prominent area of research focused on Gaussian processes mixtures, in which local components are modeled by separate \acrshort{GP}s \cite{meeds2006alternative, yuan2009variational}. Given that a single Gaussian process is an excellent approximator of nonlinear trends, the motivation for constructing such experts is not to improve the quality of approximation but rather to reduce the computational complexity and memory requirements of \acrshort{GP}s by associating every component only with a slice of the overall data, thus limiting the negative effects of cubic scaling. While this motivation is understandable from a computational standpoint, it is often unclear how to decompose a nonlinear function into a mixture of \emph{nonlinear} experts, especially when each expert has the representational flexibility of a \acrshort{GP}. In contrast, models like \gls{ILR} and \gls{HILR} rely on locally linear experts that offer a natural unit of local approximation.

Finally, the infinite mixtures used in \gls{ILR} and \gls{HILR} to regularize model complexity are rooted in Bayesian nonparametrics. Here we reference influential work on \gls{MCMC} techniques for Bayesian nonparametric density estimation \cite{escobar1995bayesian, neal2000markov, ishwaran2001gibbs, rasmussen1999infinite}, which developed the first seeds of Bayesian inference for Dirichlet processes \cite{blackwell1973ferguson, ferguson1973bayesian}. These concepts inspired comparable infinite mixture regression models. However, these attempts exclusively relied on expensive Gibbs sampling algorithms \cite{mueller1996bayesian, shababa2009nonlinear, hannah2011dirichlet, gadd2020enriched}. Instead, we focus on developing efficient deterministic \gls{VI} algorithms that dramatically improve the practical aspects of training and deploying Bayesian finite and infinite mixture models \cite{attias2000variational, blei2006variational}. For a detailed discussion on the scalability of different paradigms of Bayesian inference, we refer the reader to \cite{angelino2016patterns}.

This paper is organized as follows. In Section~\ref{sec:prelim}, we review Bayesian linear regression, finite and infinite mixture models, and variational inference. Section~\ref{sec:ilr} presents the \acrfull{ILR} model and a variational Bayes inference scheme \cite{abdulsamad2020variational}. In Section~\ref{sec:hilr}, we introduce the \acrfull{HILR}, an extension of \gls{ILR}, that incorporates multi-modal activations and parameter sharing. Finally, in Section~\ref{sec:results}, we test \gls{ILR} and \gls{HILR} on a set of simulated and real-world benchmarks.

\section{Preliminaries}
\label{sec:prelim}
This section introduces related concepts, such as Bayesian linear regression, Bayesian mixture models, Dirichlet processes, and variational inference.

\subsection{Bayesian Linear Regression}
We start by discussing the Bayesian treatment of a single component of a Bayesian local regression model, namely Bayesian linear regression \cite{minka2000bayesian}. The conditional data likelihood takes a feature vector $\vec{x} \in \mathds{R}^{m}$ as a random input variable and returns a random response variable $\vec{y} \in \mathds{R}^{d}$ according to a linear mapping $\mat{A}: \mathds{R}^{m} \to \mathds{R}^{d}$, a bias vector $\vec{c}$, and additive zero-mean noise with a precision matrix $\mat{V}$
\begin{equation*}
	\vec{y} = \mat{A} \vec{x} + \vec{c} + \vec{e}, \quad \vec{e} \sim \N(\vec{0}, \mat{V}).
\end{equation*}
For a fully Bayesian treatment, we consider all parameters of this model to be random variables on which we place proper conjugate priors. In this case, we place \gls{MN} and \gls{NW} priors on the matrix $\mat{A}$, the bias coefficient $\vec{c}$, and precision matrix $\mat{V}$
\begin{equation*}
	p(\mat{A}, \vec{c}, \mat{V}) =\MN(\mat{A} \,|\,\mat{M}, \mat{V}, \mat{K}) \N(\vec{c} \,|\,\vec{\theta}, \rho \mat{V}) \W(\mat{V} \,|\,\mat{\Phi}, \eta),
\end{equation*}
where $\mat{M}$, the mean of $\mat{A}$, is a $d \times m$ matrix and $\mat{V}$ and $\mat{K}$ are $d \times d$ and $m \times m$ that serve as row and column precision matrices of $\mat{A}$, respectively. The mean $\vec{\theta}$ is an $m$-dimensional vector, and the scalar $\rho$ modulates the amplitude of the precision. Finally, the parameters of the Wishart distribution are the $d \times d$ positive definite scale matrix $\mat{\Phi}$ and the degrees of freedom $\eta$. Due to the conjugate nature of the priors, the joint posteriors $p(\mat{A}, \vec{c}, \mat{V} \,|\,\mathcal{D})$ are matrix-normal and normal-Wishart distributions, conditioned on the data of $N$ independent and identically distributed data pairs $\mathcal{D}=\{(\vec{x}_{1}, \vec{y}_{1}), \ldots, (\vec{x}_{N}, \vec{y}_{N}) \}$.

\subsection{Bayesian Finite Mixture Models}
\label{sec:finite}
\glspl{GMM} are hierarchical latent variable models with universal approximation capabilities for arbitrary continuous densities. This insight is of central interest when connected to density estimation for local regression models, which are themselves universal nonlinear function approximators \cite{wasserman2006all}. A finite $K$-component Gaussian mixture of a random variable $\vec{x}$ is a weighted linear combination of densities
\begin{equation*}
	\vspace{-0.1cm}
	p(\vec{x} \,|\,\vec{\theta}) = \sum_{k=1}^{K} p(\vec{z} = k \,|\,\vec{\pi}) p(\vec{x} \,|\, \vec{\theta}_{k}) = \sum_{k=1}^{K} \pi_{k} \N(\vec{x} \,|\,\vec{\mu}_{k}, \mat{\Lambda}_{k}),
\end{equation*}
with $K$ unique mean vectors $\vec{\mu}_{k}$ and precision matrices $\mat{\Lambda}_{k}$. The latent quantity $\vec{z}$ is a one-hot random variable distributed according to a categorical distribution $p(\vec{z}) = \Cat(\vec{\pi})$, governed by the weights $\vec{\pi} = \{\pi_{1}, \dots, \pi_{K}\}$ that satisfy $0 \leq \pi_{k} \leq 1$ and $\sum_{k=1}^{K} \pi_{k}=1$.

The Bayesian extension of this model \cite{attias2000variational} introduces a conjugate normal-Wishart prior on the means and precision matrices $(\vec{\mu}_{k}, \mat{\Lambda}_{k}) \sim \NW(\vec{\lambda})$, where $\vec{\lambda}$ contains the hyperparameters. Furthermore, a conjugate Dirichlet prior, with a concentration parameter $\alpha$, is placed on the mixing weights $\vec{\pi} \sim \Dir(\vec{\alpha})$. This Bayesian perspective has proven effective in regularizing the shortcomings of \glspl{GMM} by allowing superfluous components to fall back onto their priors instead of severely over-fitting to small clusters. This effect can be understood as sparsification bias over $K$ \cite{beal2006variational, rousseau2011asymptotic}.

\subsection{Dirichlet Process and Stick-Breaking}
Scaling the finite mixtures in Section~\ref{sec:finite} to infinite components requires placing a nonparametric \gls{DP} prior on the parameters of the mixture. A \gls{DP} is a distribution over probability measures $\op{G}$. We write $\op{G} \sim \op{DP}(\alpha, \op{H})$, where $\alpha$ is the concentration parameter and $\op{H}$ is the base measure \cite{murphy2012machine, teh2010dirichlet}. Intuitively, a Dirichlet process is a distribution over distributions, meaning each draw $\op{G}$ is itself a distribution. The base distribution $\op{H}$ is the mean of the \gls{DP}, and the concentration parameter $\alpha$ can be interpreted as an inverse variance. The larger $\alpha$ is, the smaller the variance, and the process concentrates more of its mass around the mean distribution $\op{H}$.

We will rely on the stick-breaking construction \cite{sethuraman1994constructive} of a \gls{DP} as an algorithmic realization. Stick-breaking delivers an infinite sequence of mixture weights $\pi_{k}$ of an infinite mixture model from the stochastic process
\begin{equation*}
	\pi_{k} = s_{k} \prod_{l=1}^{k-1}\left(1 - s_{l} \right),  \quad s_{k} \sim \Beta(1, \alpha).
\end{equation*}
This process is sometimes denoted as $\vec{\pi} \sim \op{GEM}(\alpha)$ \cite{murphy2012machine}. The stick-breaking procedure describes how the random variables $s_k$, representing stick lengths, are drawn from a beta distribution and combined to obtain the mixture weights $\pi_k$. If the concentration parameter $\alpha$ increases, the magnitude of the mixing weights $\pi_k$ decreases on average, and the number of possible active components increases.

This representation of \glspl{DP} can be used to replace the priors placed on the finite Gaussian mixture model \cite{blei2006variational}. In such a setting, the base $\op{H}$ is a normal-Wishart distribution, and the sampled measure $\op{G} \sim \op{DP}(\alpha, \NW)$ is a draw of an unbounded number of parameters $(\vec{\mu}_{k}, \mat{\Lambda}_{k}) \sim \NW(\vec{\lambda})$ for an infinite number of clusters, associated with an infinite number of weights $\pi_{k}$ generated by the stick-breaking process. These draws from a Dirichlet process are discrete with probability one, which leads to the clustering effect of the \gls{DP}. Eventually, the same parameters will be sampled over and over, forcing the associated data points to cluster.

\subsection{Variational Inference of Structured Models}
\label{sec:vbem}
\gls{MCMC} \cite{brooks2011handbook}, and \gls{VI} \cite{blei2017variational} have become the two main approaches for (approximate) probabilistic inference in graphical models. While \gls{MCMC} constructs a stochastic sampling process that converges to the posterior, \gls{VI} formulates the inference task as a deterministic optimization problem. Despite its reliance on a coarser functional posterior approximation, \gls{VI} is often preferred in settings with large number of parameters. Moreover, the deterministic nature of \gls{VI} circumvents the issue of label switching in Monte-Carlo-based posterior inference of mixture models.

In a nutshell, in variational inference, a typically intractable posterior is approximated by a tractable functional distribution $q(\vec{\beta})$ that minimizes the \gls{KL} to true posterior $p(\vec{\beta} \,|\,\mathcal{D})$
\begin{equation*}
	q^{*}(\vec{\beta}) = \argmin \quad \kl (q(\vec{\beta}) \, || \, p(\vec{\beta} \,|\,\mathcal{D))}, %
\end{equation*}
where $\mathcal{D}$ is observed data and the vector $\vec{\beta}$ subsumes both the parameters $\vec{\theta}$ and hidden indicators $\vec{z}$ in structured models. Note that the \gls{KL} in the above objective is mode-seeking, meaning it will lock on one mode of a possibly multi-modal posterior. In general, the posterior $p(\vec{\beta} \,|\,\mathcal{D})$ is unknown, because the normalizer $p(\mathcal{D})$ is not tractable. In consequence, the \gls{KL} cannot be minimized directly but rather optimized via a related objective that is equal up to the constant term equivalent to the evidence $p(\mathcal{D})$
\begin{align*}
	\kl (q \, || \, p) & = \mathbb{E}_{q} \left[ \log q(\vec{\beta} )\right] - \mathbb{E}_{q} \left [ \log p(\vec{\beta} \,|\,\mathcal{D}) \right] \\
	& = \mathbb{E}_{q} \left[ \log q(\vec{\beta}) \right] - \mathbb{E}_{q} \left[ \log p(\mathcal{D}, \vec{\beta}) \right] + p(\mathcal{D}) \\
	& = \mathbb{E}_{q} \left[ \log q(\vec{\beta}) \right] - \mathbb{E}_{q} \left[ \log p(\mathcal{D}, \vec{\beta}) \right] + \mathrm{const}.
\end{align*}
This modified objective is denoted as the negative \gls{ELBO} and can be reformulated to take the traditional maximization objective of \gls{VI} algorithms
\begin{align*}
	\mathrm{ELBO}(q) & =  \mathbb{E}_{q} \left[ \log p(\mathcal{D}, \vec{\beta}) \right] - \mathbb{E}_{q} \left[ \log q(\vec{\beta}) \right] \nonumber \\
	& = \mathbb{E}_{q} \left[ \log p(\mathcal{D} \,|\,\vec{\beta}) \right] + \mathbb{E}_{q} \left[ \log p(\vec{\beta}) \right] - \mathbb{E}_{q} \left[ \log q(\vec{\beta}) \right] \nonumber \\
	& = \mathbb{E}_{q} \left[ \log p(\mathcal{D} \,|\,\vec{\beta}) \right] - \kl (q(\vec{\beta}) \, || \, p(\vec{\beta})). %
\end{align*}

The choice of the approximate posterior distribution $q(\vec{\beta})$ is open. In this paper, we focus on variational Bayes methods that rely on the (structured) mean-field assumption as a general recipe for maximizing the \gls{ELBO} \cite{opper2001advanced, beal2003variational}. This approximation requires that the posterior factorizes over the set of the hidden variables $q(\vec{\beta}) = \prod_{i=1}^{M} q_{i}(\beta_{i})$. It is emphasized that no other assumptions are made about $q(\vec{\beta})$. The resulting posterior will be determined solely by the assumed likelihood and priors. 

Specifically, we follow the scheme of \gls{VBEM} as a probabilistic generalization of \gls{EM}. This approach constitutes a coordinate ascent scheme that iteratively optimizes the \gls{ELBO} for individual factors of the approximate posterior $q(\vec{\beta})$ while holding the others constant
\begin{equation*}
	\ln q_{j}(\beta_{j}) = \mathbb{E}_{q_{i \neq j}} \left[ \log p(\mathcal{D}, \vec{\beta}) \right] + \mathrm{const}.
\end{equation*}
A more practical version of this optimization can be achieved by using \gls{SVI}, a batched stochastic gradient ascent approach. In the case of the conjugate exponential family, \gls{SVI} not only facilitates scalability over large datasets but also resembles a natural gradient ascent algorithm on the \gls{ELBO} with favorable convergence properties \cite{hoffman2013stochastic}.

\begin{figure*}[!t]
	\centering
	\includeinkscape[width=400pt, svgpath=figures/plates/]{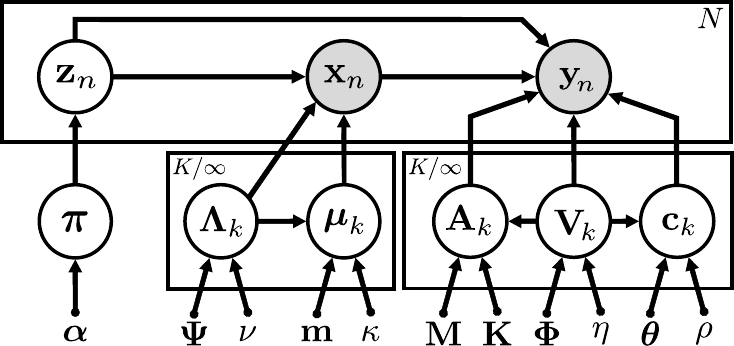}
	\caption{A unified plate notation for infinite mixtures of Bayesian local regression. Assuming Gaussian and linear Gaussian densities, the basis parameters $(\vec{\mu}_{k}, \mat{\Lambda}_{k})$ are sampled from a normal-Wishart distribution, while the regression parameters $(\mat{A}_{k}$, $\vec{c}_{k})$ and precision matrices $\mat{V}_{k}$ are sampled from a matrix-Normal-Wishart, for every component $k$. The latent variables $\vec{z}_{n}$ assign every $\vec{x}_{n}$ and $\vec{y}_{n}$ to a component and are drawn from a categorical distribution parameterized by $\vec{\pi}$. The mixture weights $\vec{\pi}$ are generated by a stick-breaking process with a concentration parameter $\alpha$.}
	\label{fig:ilr_plates}
	\vspace{-0.25cm}
\end{figure*}

\section{The Infinite Mixture of Local Linear Regression Model}
\label{sec:ilr}
Using the previously presented concepts of Bayesian linear regression, Bayesian mixture models, and Dirichlet processes, we now construct the Bayesian \acf{ILR} model. 

Our approach to tackling the regression problem is a Bayesian joint density estimation technique over the input $\vec{x}$ and target $\vec{y}$, both conditioned on a latent discrete label $\vec{z}$. To avoid the need for a fixed number of components over $\vec{z}$, we assume a generative model driven by a stick-breaking process with a concentration parameter $\alpha$ and a base distribution $\op{H}$, which is a product of a normal-Wishart, a Matrix-normal, and Wishart distribution.

The data generation proceeds as depicted in Figure~\ref{fig:ilr_plates}. The stick-breaking process is sampled to generate the categorical weights $\vec{\pi}$, mixture activations $\{ \vec{\mu}_{k}, \mat{\Lambda}_{k} \}_{k=1}^{\infty}$, and regression parameters $\{ \mat{A}_{k}, \vec{c}_{k}, \mat{V}_{k} \}_{k=1}^{\infty}$
\begin{equation*}
	\begin{gathered}
		\vec{\pi}(\vec{s}) \sim \op{GEM}(\alpha), \\
		\mat{\Lambda}_{k} \sim \W(\mat{\Psi}, \nu),  \,\, \vec{\mu}_{k} \sim \N(\vec{m}, \kappa \mat{\Lambda}_{k}), \\
		\mat{V}_{k} \sim \W(\mat{\Phi}, \eta),  \,\, \mat{A}_{k} \sim \MN(\mat{M}, \mat{K}, \mat{V}_{k}),  \,\, \vec{c}_{k} \sim \N(\vec{\theta}, \rho \mat{V}_{k}),
	\end{gathered}
\end{equation*}
and those parameters are used to generate the one-hot labels $\vec{z}_{n}$ and input-output data pairs $\vec{x}_{n}, \vec{y}_{n}$
\begin{gather*}
	\vec{z}_{n} \sim \Cat(\vec{\pi}(\vec{s})), \, \, \vec{x}_{n} \sim \N(\vec{\mu}_{z_{n}}, \mat{\Lambda}_{z_{n}}), \\
	\vec{y}_{n} \sim \N(\mat{A}_{z_{n}} \vec{x}_{n} + \vec{c}_{z_{n}}, \mat{V}_{z_{n}}).
\end{gather*}
Notice that the densities over the input space, parameterized by $(\vec{\mu}_{k}, \mat{\Lambda}_{k})$, naturally play the role of basis functions or so-called receptive fields as in the receptive field weighted regression (RFWR) \cite{schaal2002scalable} and locally weighted projection regression (LWPR) \cite{vijayakumar2005incremental} algorithms.

Given these modeling assumptions, the regression objective can be cast as a joint density inference problem of the posterior $p(\mat{Z}, \vec{s}, \vec{\mu}, \allowbreak \mat{\Lambda}, \mat{A}, \vec{c}, \mat{V} \,|\,\mat{X}, \mat{Y})$, where $\mat{X} = \{\vec{x}_{1}, \dots, \vec{x}_{N} \}$, $\mat{Y} = \{\vec{y}_{1}, \dots, \vec{y}_{N} \}$, and $\mat{Z} = \{\vec{z}_{1}, \dots, \vec{z}_{N} \}$. The result of this inference is then leveraged to perform predictions via conditioning and marginalization.

\subsection{Variational Bayes for \gls{ILR}}
For inference, we focus on a variational Bayes \gls{EM} algorithm that alternates between a variational expectation step (E-step) and a maximization step (M-step). Deriving such an algorithm for this model requires pinning down the following definitions of the likelihood, prior, and posterior.

\textbf{Complete Data Likelihood.} For the general case of multivariate regression with $m$ inputs and $d$ outputs, we assume the following structured joint likelihood over the inputs, outputs, and indicator variables
\begin{align*}
	p(\mat{X}, \mat{Y}, \mat{Z} \,|\,.) & =  p(\mat{Z}) \, p(\mat{X} \,|\,\mat{Z}) \, p(\mat{Y} \,|\,\mat{X}, \mat{Z}) \\
	& =
	\begin{aligned}[t]
		& \prod_{n=1}^{N} \Cat(\vec{z}_{n} \,|\,\vec{\pi}(\vec{s})) \\
		& \times \prod_{n=1}^{N} \prod_{k=1}^{\infty} \N(\vec{x}_{n} \,|\,\vec{\mu}_{k}, \mat{\Lambda}_{k})^{z_{nk}} \\
		& \times \prod_{n=1}^{N} \prod_{k=1}^{\infty} \N(\vec{y}_{n} \,|\,\mat{A}_{k} \vec{x}_{n} + \vec{c}_{k}, \mat{V}_{k})^{z_{nk}},
	\end{aligned}
\end{align*}
where the dimensions of all quantities follow the notation of Bayesian linear regression.

\textbf{Infinite Conjugate Prior.} We construct a factorized conjugate infinite mixture prior
\begin{equation*}
	p(\vec{s}, \vec{\mu}, \mat{\Lambda}, \mat{A}, \vec{c}, \mat{V}) = 
	\begin{aligned}[t]
		& p(\vec{s}) \, p(\vec{\mu}| \mat{\Lambda}) \, p(\mat{\Lambda}) \\
		& \times p(\mat{A}| \mat{V}) \, p(\vec{c}| \mat{V}) \, p(\mat{V}).
	\end{aligned}
\end{equation*}
This prior assumes that the cluster means $\vec{\mu}_{k}$ and precision matrices $\mat{\Lambda}_{k}$ are sampled from normal-Wishart distributions
\begin{equation*}	 
	p(\vec{\mu} \,|\,\mat{\Lambda}) \, p(\mat{\Lambda}) = \prod_{k=1}^{\infty} \N(\vec{\mu}_{k} \,|\,\vec{m}_{0}, \kappa_{0} \mat{\Lambda}_{k}) \W(\mat{\Lambda}_{k} \,|\,\mat{\Psi}_{0}, \nu_{0}),
\end{equation*}
while matrix-normal-Wishart and normal-Wishart priors are placed on the regression coefficients $(\mat{A}_{k}, \vec{c}_{k})$ and the precision matrices $\mat{V}_k$
\begin{align*}	 
	p(\mat{A} \,|\,\mat{V}) p(\vec{c} \,|\,\mat{V}) p(\mat{V}) = \prod_{k=1}^{\infty} & \MN(\mat{A}_{k} \,|\,\mat{M}_{0}, \mat{K}_{0}, \mat{V}_{k}) \\
	& \hspace{-0.5cm} \times \N(\vec{c}_{k} \,|\,\vec{\theta}_{0}, \rho_{0} \mat{V}_{k}) \W(\mat{V}_{k} \,|\,\mat{\Phi}_{0}, \eta_{0}).
\end{align*}
The parameters $\pi_k$ are generated by a stick-breaking process $\pi_{k}(\vec{s}) = s_{k} \prod_{l=1}^{k-1} (1 - s_{l})$, where the stick lengths $\vec{s}$ are independently beta distributed
\begin{equation*}
	p(\vec{s}) = \prod_{k=1}^{\infty} \Beta(s_{k} \,|\,1, \alpha_{0}).
\end{equation*}

\textbf{Truncated Mean-Field Posterior.} We rely on a structured mean-field approximation of the posterior \cite{opper2001advanced} that factorizes between the labels $q(\mat{Z})$ and the remaining parameters $q(\vec{s}, \vec{\mu}, \mat{\Lambda}, \mat{A}, \vec{c}, \mat{V})$, thus automatically leading to the decomposition $p(. \,|\,\mathcal{D}) \approx q(\mat{Z}) \, q(\vec{s}) \, q(\vec{\mu}, \mat{\Lambda}) \, q(\mat{A}, \vec{c}, \mat{V})$.
Further, we follow \cite{blei2006variational} by allowing a truncation of the posterior while maintaining an infinite prior, so that $q(s_{\scriptscriptstyle K}\!=\!1) = 1$, implying that $\pi_{k} = 0$ for $k > K$ 
\begin{equation*}
	p(. \,|\,\mathcal{D}) \approx
	\begin{aligned}[t]
		& \prod_{n=1}^{N} \Cat(\vec{z}_{n} \,|\,\vec{r}_{n}) \prod_{k=1}^{K-1} \Beta(s_{k} \,|\,\gamma_{k}, \alpha_{k}) \\
		& \times \prod_{k=1}^{K} \N(\vec{\mu}_{k} \,|\,\vec{m}_{k}, \kappa_{k} \mat{\Lambda}_{k}) \W(\mat{\Lambda}_{k} \,|\,\mat{\Psi}_{k}, \nu_{k}) \\
		& \times 
		\begin{aligned}[t]
			\prod_{k=1}^{K} & \MN(\mat{A}_{k} \,|\,\mat{M}_{k}, \mat{K}_{k}, \mat{V}_{k}) \\
			& \times \N(\vec{c}_{k} \,|\,\vec{\theta}_{k}, \rho_{k} \mat{V}_{k}) \W(\mat{V}_{k} \,|\,\mat{\Phi}_{k}, \eta_{k}),
		\end{aligned}
	\end{aligned}
\end{equation*}
where $\vec{r}_{n}$ are the expected responsibilities of the mixture. During evaluation, the truncation threshold $K$ is chosen to be very high and is seldom reached. %

\textbf{Variational Expectation Step.} In the E-step, the responsibilities are computed by following the recipe of \gls{VBEM} in Section~\ref{sec:vbem}. The responsibilities are the variational parameters of the posterior categorical
\begin{align*}
	\log q(\mat{Z}) & = 
	\begin{aligned}[t]
		& \mathbb{E}_{q(\vec{s})} \left[ \log p(\mat{Z} \,|\,\vec{\pi}(\vec{s})) \right] + \mathbb{E}_{q(\vec{\mu}, \mat{\Lambda})} \left[ \log p(\mat{X} \,|\,\mat{Z}) \right] \\
		& + \mathbb{E}_{q(\mat{A}, \vec{c}, \mat{V})} \left[ \log p(\mat{Y} \,|\,\mat{X},  \mat{Z}) \right] + \const
	\end{aligned} \\
	& = \sum_{n=1}^{N} \sum_{k=1}^{K} z_{nk} \log r_{nk}.
\end{align*}
The expectations associated with the data likelihoods of $\vec{x}_{n}$ and $\vec{y}_{n}$ can be straightforwardly computed \cite{bishop2006pattern}. However, the expectations associated with infinite-dimensional categorical require more careful consideration \cite{blei2006variational}. We provide the necessary details in the appendix.

\textbf{Variational Maximization Step.} The M-step updates the remaining variational distributions given an approximation of the categorical posterior as follows
\begin{align*}
	\log q(\vec{s}) & = \mathbb{E}_{q(\mat{Z})} \left[ \log  p(\mat{Z} \,|\,\vec{\pi}(\vec{s})) \right] + \log p(\vec{s}) + \const, \\
	\log q(\vec{\mu}, \mat{\Lambda}) & = \mathbb{E}_{q(\mat{Z})} \left[ \log p(\mat{X} \,|\,\mat{Z}) \right] + \log p(\vec{\mu}, \mat{\Lambda}) + \const, \\
	\log q(\mat{A}, \vec{c}, \mat{V}) & = 
	\begin{aligned}[t]
		& \mathbb{E}_{q(\mat{Z})} \left[ \log p(\mat{Y} \,|\,\mat{X}, \mat{Z}) \right] \\
		& + \log p(\mat{A}, \vec{c}, \mat{V}) + \const.
	\end{aligned}
\end{align*}
Each of these updates reflects a conjugate computation of $K$ log-posterior densities given a log-prior and a log-likelihood weighted by the responsibilities $r_{nk} = \mathbb{E} [z_{nk}]$. We provide more details and general computational recipes based on conjugacy rules in the appendix. %

\subsection{Posterior Predictive Distribution}
\label{sec:ilr_predictive}
For predicting the function value $\hat{\vec{y}}$ conditioned on a test query $\hat{\vec{x}}$, we marginalize the likelihood over the posterior parameters to get the joint posterior predictive density. To make the marginalization tractable, we replace the true posterior with our approximate variational posterior $q(. \,|\,\mathcal{D})$ inferred under a training dataset $\mathcal{D}$. The conditional predictive for a single component $\vec{z}=k$ is a conditional multivariate Student's t-distribution of the form
\begin{align*}
	p(\hvec{y} \,|\,\hvec{x}, \hvec{z}\!=\!k, \mathcal{D}) & = \mathbb{E}_{q(\mat{A}, \vec{c}, \mat{V})} \left[ p(\hvec{y} \,|\,\hvec{x}, \mat{A}_{k}, \vec{c}_{k}, \mat{V}_{k}) \right] \\
	& = \op{T} \left( \mat{M}_{k} \hvec{x} + \vec{\theta}_{k}, a_{k} \mat{\Phi}_{k}, \eta_{k} + 1 \right),
\end{align*}
where we have defined
\begin{equation*}
	a_{k} = 1 - \vec{u}^{\top} \left( \mat{L}_{k} + \vec{u} \vec{u}^{\top} \right)^{-1} \vec{u},
\end{equation*}
with $\vec{u} = [\hvec{x}, \, 1]^{\top}$ and $\mat{L}_{k} = \op{Block} \left(\mat{K}_{k}, \rho_{k} \right)$.

Additionally, the joint activation of a component $k$ is a Student's t-distribution weighted by the expected categorical probability under the posterior stick-breaking process
\begin{align*}
	p(\hvec{x}, \hvec{z}\!=\!k \,|\,\mathcal{D}) & \propto \mathbb{E}_{q(\vec{s})} \left[ p(\hvec{z}\!=\!k \,|\, \vec{\pi}(\vec{s})) \right] \mathbb{E}_{q(\vec{\mu}, \mat{\Lambda})} \left[ p(\hvec{x} \,|\,\vec{\mu}_{k}, \mat{\Lambda}_{k}) \right] \\
	& = 
	\begin{aligned}[t]
		& \frac{\gamma_{k}}{\gamma_{k} + \alpha_{k}} \prod_{l=1}^{k-1} \left( 1 - \frac{\gamma_{l}}{\gamma_{l} + \alpha_{l}} \right) \\
		& \times \op{T} \left( \vec{\mu}_{k}, \frac{\kappa_{k}}{1 + \kappa_{k}} \mat{\Psi}_{k}, \nu_{k} + 1 \right).
	\end{aligned}
\end{align*}
These $K$-activation probabilities enable two prediction techniques. A \textit{mode-prediction}, where the most likely active component is selected and used to perform prediction with the corresponding linear regression model, or a \textit{mean-prediction}, that averages the predictions of all components weighted by their activation probabilities.

\subsection{Computational Complexity}
We calculate the training-time computational cost to be $\mathcal{O}(NK(d+m)^3)$, which can be straightforwardly reduced to $\mathcal{O}(LK(d+m)^3)$ by applying stochastic updates \cite{hoffman2013stochastic}, where $L$ is the batch size. This result shows linear scalability with the data, which is considerably more efficient than simple variants of \gls{GPR}. The test-time complexity of a mean prediction is $\mathcal{O}(K(d^3+dm))$, which combines the input membership query and the linear matrix transformation for every model $k$. This computation is, in contrast to \gls{GPR}, independent of the training data size, hence the advantage of memoryless locally-parametric representations during real-time critical applications.

\begin{figure*}[t!]
	\centering
	\includeinkscape[width=450pt, svgpath=figures/plates/]{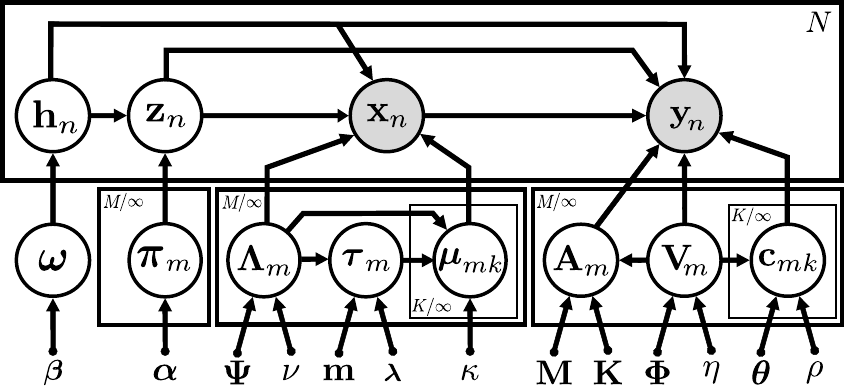}
	\caption{A unified plate notation for hierarchical infinite tied mixtures of Bayesian local regression models. This model outlines a two-level architecture that allows sharing of parameters between single components in order to compress the representation. It can be interpreted as a local regression model with multi-modal activations. Each unit of the upper level is itself a mixture of local regression models that share the same slope $\mat{A}_{m}$ and output precision $\mat{V}_{m}$. Each of these $m$ different slopes can be activated at $k$ unique lower-level input regions centered around $\vec{\mu}_{mk}$ and tied via a shared input precision $\mat{\Lambda}_{m}$. Both the upper- and lower-level mixtures are governed by independent Dirichlet process priors.}
	\label{fig:hilr_plates}
	\vspace{-0.25cm}
\end{figure*}

\section{The Hierarchical Infinite Mixture of Local Linear Regression Model}
\label{sec:hilr}
The local regression model presented in Section~\ref{sec:ilr} offers a very flexible and well-regularized alternative to previously developed approaches \cite{schaal1998constructive, vijayakumar2005incremental, meier2014incremental}. However, like similar local regression representations, it cannot account for shift-invariance across the input space. This drawback can cause it to generate duplicate regression components to account for similar local function trends across disconnected regions of the input space. This limitation results from the hierarchical design that directly couples activations and local function units via a one-to-one correspondence and enforces a uni-modal activation per regression component. This coupling can be observed in the definition of the likelihood in Section~\ref{sec:ilr}, where the activation and the local regression units share the same assignment variable. 

It stands to reason that \gls{ILR} does not offer enough flexibility and hinders parameter sharing between components. Therefore, we present a modified formulation of \gls{ILR} that explicitly accounts for shift-invariance in the input space and provides the freedom to create regression units with multi-modal, theoretically infinitely-modal, activations, if needed. The resulting model, \acf{HILR}, is an infinite mixture over infinite mixtures that shares similarities with existing representations developed for hierarchical clustering \cite{yerebakan2014infinite, nguyen2014bayesian, huynh2016scalable}. 

We start by describing the generative process of the hierarchical mixture over mixtures as depicted in Figure~\ref{fig:hilr_plates}. \gls{HILR} consists of two levels. At the upper level, a meta Dirichlet process generates the stick-breaking weights $\vec{\omega}$, the meta-activations $\{ \vec{\tau}_{m}, \mat{\Lambda}_{m} \}_{m=1}^{\infty}$, and the shared slope and output precision matrices $\{ \mat{A}_{m}, \mat{V}_{m} \}_{m=1}^{\infty}$
\begin{equation*}
	\begin{gathered}
		\vec{\omega}(\vec{t}) \sim \op{GEM}(\beta), \\
		\mat{\Lambda}_{m} \sim \W(\mat{\Psi}, \nu), \,\, \vec{\tau}_{m} \sim \N(\vec{m}, \lambda \mat{\Lambda}_{m}), \\
		\mat{V}_{m} \sim \W(\mat{\Phi}, \eta),  \,\, \mat{A}_{m} \sim \MN(\mat{M}, \mat{K}, \mat{V}_{m}),
	\end{gathered}
\end{equation*}
where $\beta$ is the concentration parameter of the upper-level \gls{DP} and $\vec{t}$ are the generated stick lengths. 

\gls{HILR}'s ability to account for shift-invariance is a result of allowing multi-modal activations. That way, every regression unit can be associated with multiple local regions in the input space. This structure is realized by endowing every upper-level component $m$ with its own local \gls{DP}. These lower-level \glspl{DP}, with a concentration parameter $\alpha$, generate the weights $\vec{\pi}_{m}$, the activation centers $\{ \vec{\mu}_{mk} \}_{k=1}^{\infty}$ and the shift coefficients $\{ \vec{c}_{mk} \}_{k=1}^{\infty}$
\begin{gather*}
	\vec{\pi}_{m} (\vec{s}_{m}) \sim \op{GEM}(\alpha), \\ 
	\vec{\mu}_{mk} \sim \N(\vec{\tau}_{m}, \kappa \mat{\Lambda}_{m}), \, \, \vec{c}_{mk} \sim \N(\vec{\theta}, \rho \mat{V}_{m}).
\end{gather*}
We assume here that the lower-level centers $\vec{\mu}_{mk}$ and coefficients $\vec{c}_{mk}$ are tied via their respective upper-level precision matrices $\mat{\Lambda}_{m}$ and $\mat{V}_{m}$. 

Finally, given all necessary parameters, the upper- and lower-level labels $\vec{h}_{n}, \vec{z}_{n}$ and the data pairs $\vec{x}_{n}, \vec{y}_{n}$ are sampled according to the likelihoods
\begin{gather*}
	\vec{h}_{n} \sim \Cat(\vec{\omega}(\vec{t})), \, \, \vec{z}_{n} \sim \Cat(\vec{\pi}(\vec{s}), \vec{h}_{n}), \\
	\vec{x}_{n} \sim \N(\vec{\mu}_{h_{n}, z_{n}}, \mat{\Lambda}_{h_{n}}), \, \, \vec{y}_{n} \sim \N(\mat{A}_{h_{n}} \vec{x}_{n} + \vec{c}_{h_{n}, z_{n}}, \mat{V}_{h_{n}}).
\end{gather*}

Notice how the upper-level generative process resembles the structure of \gls{ILR}. However, in this model, the meta-activations  $(\vec{\tau}_{m}, \mat{\Lambda}_{m})$ induce another hierarchy over the lower-level multi-modal activation centers $\vec{\mu}_{mk}$. Moreover, the upper-level \gls{DP} only accounts for the shared slope of the local regression units $\mat{A}_{m}$, whereas the shift coefficients $\mat{c}_{mk}$ are induced at the lower level, as they are influenced by the individual activation centers $\vec{\mu}_{mk}$.

Similar to the procedure in Section~\ref{sec:ilr}, we aim to infer a posterior over the parameters of this model given a set of inputs $\mat{X} = \{\vec{x}_{1}, \dots, \vec{x}_{N} \}$ and outputs $\mat{Y} = \{\vec{y}_{1}, \dots, \vec{y}_{N} \}$. To that end, we derive a structured variational Bayes expectation-maximization algorithm for this model. We start the derivation by defining the complete data likelihood, the infinite prior, and the mean-field posterior factorization.

\textbf{Complete Data Likelihood.} The likelihood model is a two-level precision-tied joint density over the observations $(\mat{X}, \mat{Y})$ and the one-hot upper- and lower-labels $(\mat{H}, \mat{Z})$
\begin{align*}
	p(.) & =  p(\mat{H}) \, p(\mat{Z} \,|\,\mat{H}) \, p(\mat{X} \,|\,\mat{H}, \mat{Z}) \, p(\mat{Y} \,|\,\mat{H}, \mat{Z}, \mat{X}) \\
	& =
	\begin{aligned}[t]
		& \prod_{n=1}^{N} \Cat(\vec{h}_{n} \,|\,\vec{\omega}(\vec{t})) \\
		& \times \prod_{n=1}^{N} \Cat(\vec{z}_{n} \,|\,\vec{\pi}(\vec{s}), \vec{h}_{n}) \\
		& \times \prod_{n=1}^{N} \prod_{m=1}^{\infty} \prod_{k=1}^{\infty} \N(\vec{x}_{n} \,|\,\vec{\mu}_{mk}, \mat{\Lambda}_{m})^{z_{nk} \times h_{nm}} \\
		& \times \prod_{n=1}^{N} \prod_{m=1}^{\infty} \prod_{k=1}^{\infty} \N(\vec{y}_{n} \,|\,\mat{A}_{m} \vec{x}_{n} + \vec{c}_{mk}, \mat{V}_{m})^{z_{nk} \times h_{nm}},
	\end{aligned}
\end{align*}
where the conditioning of the lower-level labels $\vec{z}_{n}$ on the upper-level labels $\vec{h}_{n}$ manifests as a one-hot selector of the corresponding lower-level categoricals 
\begin{align*}
	\Cat(\vec{z}_{n} \,|\,\vec{\pi}(\vec{s}), \vec{h}_{n}) & = \prod_{m=1}^{\infty} \Cat(\vec{z}_{n} \,|\,\vec{\pi}_{m}(\vec{s}_{m}))^{h_{nm}} \\
	& = \prod_{m=1}^{\infty} \prod_{k=1}^{\infty} \pi_{mk}^{z_{nk} \times h_{nm}}.
\end{align*}

\textbf{Infinite Conjugate Prior.} We assume a factorized two-level precision-tied conjugate infinite mixture prior
\begin{equation*}
	p(\vec{t}, \vec{s}, \vec{\mu},  \vec{\tau},  \mat{\Lambda}, \mat{A}, \vec{c}, \mat{V}) = 
	\begin{aligned}[t]
		& p(\vec{t}) \, p(\vec{s}) \\
		& \times p(\vec{\mu} \,|\,\vec{\tau}, \mat{\Lambda}) \, p(\vec{\tau} \,|\,\mat{\Lambda}) \, p(\mat{\Lambda}) \\
		& \times p(\mat{A} \,|\,\mat{V}) \, p(\vec{c}| \mat{V}) \, p(\mat{V}).
	\end{aligned}
\end{equation*}
The meta-activation prior is a normal-Wishart distribution over the meta-centers $\vec{\tau}_{m}$ and precision matrices $\mat{\Lambda}_{m}$
\begin{equation*}
	p(\vec{\tau} \,|\,\mat{\Lambda}) p(\mat{\Lambda}) = \prod_{m=1}^{\infty} \N(\vec{\tau}_{m} \,|\,\vec{m}_{0}, \lambda_{0} \mat{\Lambda}_{m}) \W(\mat{\Lambda}_{m} \,|\,\mat{\Psi}_{0}, \nu_{0}),
\end{equation*}
while the activation centers $\vec{\mu}_{mk}$ are sampled from a conditional normal distribution
\begin{equation*}
	p(\vec{\mu} \,|\,\vec{\tau}, \mat{\Lambda})= \prod_{m=1}^{\infty} \prod_{k=1}^{\infty} \N(\vec{\mu}_{mk} \,|\,\vec{\tau}_{m}, \kappa_{0} \mat{\Lambda}_{m}).
\end{equation*}
The mappings $\mat{A}_{m}$ and precision matrices $\mat{V}_{m}$ are sampled form a matrix-normal-Wishart
\begin{equation*}
	p(\mat{A} \,|\,\mat{V}) p(\mat{V}) = \prod_{m=1}^{\infty} \MN(\mat{A}_{m} \,|\,\mat{M}_{0}, \mat{K}_{0}, \mat{V}_{m}) \W(\mat{V}_{m} \,|\,\mat{\Phi}_{0}, \eta_{0}).
\end{equation*}
while the biases $\vec{c}_{mk}$ are drawn from a $K$-tied conditional normal distribution
\begin{equation*}
	p(\vec{c} \,|\,\mat{V}) = \prod_{m=1}^{\infty} \prod_{k=1}^{\infty} \MN(\vec{c}_{mk} \,|\,\vec{\theta}_{0}, \rho_{0} \mat{V}_{m}).
\end{equation*}
Finally, the stick-breaking priors $p(\vec{t}$, $\vec{s})$ follow the definitions from Section~\ref{sec:ilr}
\begin{align*}
	p(\vec{t}) & = \prod_{m=1}^{\infty} \Beta(t_{m} \,|\,1, \beta_{0}), \\
	p(\vec{s}) & = \prod_{m=1}^{\infty} \prod_{k=1}^{\infty} \Beta(s_{mk} \,|\,1, \alpha_{0}).
\end{align*}

\textbf{Truncated Mean-Field Posterior.} We assume a structured decomposition of the posterior that leads to conjugate computation while maintaining the dependencies between the discrete labels, the input activations, and the regression parameters, respectively. Moreover, we apply the truncation scheme from \cite{blei2006variational} to establish a tractable posterior approximation while maintaining an infinite prior
\begin{align*}
	p(. \,|\,\mathcal{D}) & \approx q(\mat{H}) \, q(\mat{Z} \,|\,\mat{H}) \, q(\vec{t}) \, q(\vec{s}) \, q(\vec{\mu}, \vec{\tau}, \mat{\Lambda}) \, q(\mat{A}, \vec{c}, \mat{V}) \\
	& \hspace{-0.5cm} =
	\begin{aligned}[t]
		& \prod_{n=1}^{N} \Cat(\vec{h}_{n} \,|\,\vec{g}_{n}) \prod_{n=1}^{N} \Cat(\vec{z}_{n} \,|\,\vec{r}_{n}, \vec{h}_{n}) \\
		& \times \prod_{m=1}^{M-1} \Beta(t_{m} \,|\,\delta_{m}, \beta_{m}) \\
		& \times \prod_{m=1}^{M-1} \prod_{k=1}^{K-1} \Beta(s_{mk} \,|\,\gamma_{mk}, \alpha_{mk}) \\
		& \times \prod_{m=1}^{M} \N(\vec{\tau}_{m} \,|\,\vec{m}_{m}, \lambda_{m} \mat{\Lambda}_{m}) \W(\mat{\Lambda}_{m} \,|\,\mat{\Psi}_{m}, \nu_{m}) \\
		& \times \prod_{m=1}^{M} \MN(\mat{A}_{m} \,|\,\mat{M}_{m}, \mat{K}_{m}, \mat{V}_{m}) \W(\mat{V}_{m} \,|\,\mat{\Phi}_{m}, \eta_{m}) \\
		& \times \prod_{m=1}^{M} \prod_{k=1}^{K} \N(\vec{\mu}_{mk} \,|\,\vec{\tau}_{m}, \kappa_{mk} \mat{\Lambda}_{m}) \N(\vec{c}_{mk} \,|\,\vec{\theta}_{mk}, \rho_{mk} \mat{V}_{k}),
	\end{aligned}
\end{align*}
where $\vec{g}_{n}$ and $\vec{r}_{n}$ are the upper- and lower-level posterior responsibilities, respectively.

\textbf{Variational Expectation Step.} The E-step computes the joint posterior categorical over joint labels $\mat{H}$ and $\mat{Z}$
\begin{align*}
	\log q(\mat{Z} \,|\,\mat{H}) & =
	\begin{aligned}[t]
		& \mathbb{E}_{q(\vec{s})} \left[ \log p(\mat{Z} \,|\,\mat{H}) \right] \\
		& + \mathbb{E}_{q(\vec{\mu}, \vec{\tau}, \mat{\Lambda})} \left[ \log p(\mat{X} \,|\,\mat{H}, \mat{Z}) \right] \\ 
		& + \mathbb{E}_{q(\mat{A}, \vec{c}, \mat{V})} \left[ \log p(\mat{Y} \,|\,\mat{H}, \mat{Z}, \mat{X}) \right] + \const
	\end{aligned} \\
	& = \sum_{n=1}^{N} \sum_{m=1}^{M} \sum_{k=1}^{K} h_{nm} z_{nmk} \log r_{nmk},
\end{align*}
\begin{align*}
	\log q(\mat{H}) & = \mathbb{E}_{q(\vec{t})} \left[ \log p(\mat{H}) \right] + \log q(\mat{Z} \,|\,\mat{H}) + \const \\
	& = \sum_{n=1}^{N} \sum_{m=1}^{M} h_{nm} \log g_{nm},
\end{align*}
where these expectations can computed in a similar fashion to Section~\ref{sec:ilr}. More details can be found in the appendix.

\textbf{Variational Maximization Step.} The M-step updates the variational gating, activations, and regression parameters
\begin{align*}
	\log q(\vec{t})                   & = \mathbb{E}_{q(\mat{H})} \left[ \log p(\mat{H}) \right] + \log p(\vec{t}) + \const, \\[1em]
	\log q(\vec{s})                   & = \mathbb{E}_{q(\mat{H}, \mat{Z})} \left[ \log p(\mat{Z} \,|\,\mat{H}) \right] + \log p(\vec{s}) + \const, \\[1em]
	\log q(\vec{\mu}) & = 
	\begin{aligned}[t]
		& \mathbb{E}_{q(\mat{H}, \mat{Z}, \vec{\tau}, \mat{\Lambda})} \left[ \log p(\mat{X} \,|\,\mat{H}, \mat{Z}) \right] \\
		& + \mathbb{E}_{q(\vec{\tau}, \mat{\Lambda})} \left[ \log p(\vec{\mu} \,|\,\vec{\tau}, \mat{\Lambda}) \right] + \const,
	\end{aligned} \\[1em]
	\log q(\vec{\tau}, \mat{\Lambda}) & = 
	\begin{aligned}[t]
		& \mathbb{E}_{q(\mat{H}, \mat{Z}, \vec{\mu})} \left[ \log p(\mat{X} \,|\,\mat{H}, \mat{Z}) \right] + \log p(\vec{\tau}, \mat{\Lambda}) \\
		& + \mathbb{E}_{q(\vec{\mu})} \left[ \log p(\vec{\mu} \,|\,\vec{\tau}, \mat{\Lambda}) \right] + \const,
	\end{aligned} \\[1em]
	\log q(\vec{c}) & = 
	\begin{aligned}[t]
		& \mathbb{E}_{q(\mat{H}, \mat{Z}, \mat{A}, \mat{V})} \left[ \log p(\mat{Y} \,|\,\mat{H}, \mat{Z}, \mat{X}) \right] \\
		& + \mathbb{E}_{q(\mat{V})} \left[ \log p(\vec{c} \,|\,\mat{V}) \right] + \const,
	\end{aligned} \\[1em]
	\log q(\mat{A}, \mat{V}) & = 
	\begin{aligned}[t]
		& \mathbb{E}_{q(\mat{H}, \mat{Z}, \vec{c})} \left[ \log p(\mat{Y} \,|\,\mat{H}, \mat{Z}, \mat{X}) \right] + \log p(\mat{A}, \mat{V})  \\
		& + \mathbb{E}_{q(\vec{c})} \left[ \log p(\vec{c} \,|\,\mat{V}) \right] + \const.
	\end{aligned}
\end{align*}
As previously stated, these updates resemble posterior computations weighted by the responsibilities $g_{nm} = \mathbb{E} [h_{nm}]$ and $r_{nmk} = \mathbb{E} [z_{nmk}]$. The appendix provides further details.

\subsection{Posterior Predictive Distribution}
\label{sec:predictive}
Prediction with \gls{HILR} is akin to that with \gls{ILR}, as described in Section~\ref{sec:ilr_predictive}. We briefly state the conditional predictive for a component $\vec{h}=m$ and an activation $\vec{z}=k$ as
\begin{align*}
	p(\hvec{y} \,|\, \hvec{x}, \hvec{h}\!=\!m, \hvec{z}\!=\!k, \mathcal{D}) & = \mathbb{E}_{q(\mat{A}, \vec{c}, \mat{V})} \left[ p(\hvec{y} \,|\, \hvec{x}, \mat{A}_{m}, \vec{c}_{mk}, \mat{V}_{m}) \right] \\
	& = \op{T} \left( \mat{M}_{m} \hvec{x} + \vec{\theta}_{mk}, a_{mk} \mat{\Phi}_{m}, \eta_{m} + 1 \right),
\end{align*}
where the computation of $a_{mk}$ is similar to that in Section~\ref{sec:ilr_predictive}. Further, the weight of the $k-$th activation of the $m-$th component is computed as follows
\begin{align*}
	p(\hvec{x}, \hvec{h}\!=\!m, \hvec{z}\!=\!k \,|\, \mathcal{D})
	& \propto
	\begin{aligned}[t]
		& \mathbb{E}_{q(\vec{t})} \left[ p(\hvec{h}\!=\!m \,|\, \vec{\omega}(\vec{t})) \right] \\
		& \times \mathbb{E}_{q(\vec{s}_{m})} \left[ p(\hvec{z}\!=\!k \,|\, \vec{\pi}_{m}(\vec{s}_{m})) \right] \\ 
		& \times \mathbb{E}_{q(\vec{\mu}, \vec{\tau}, \mat{\Lambda})} \left[ p(\hvec{x} \,|\, \vec{\mu}_{mk}, \mat{\Lambda}_{m}) \right]
	\end{aligned} \\
	& \hspace{-2.0cm} = 
	\begin{aligned}[t]
		& \frac{\delta_{m}}{\delta_{m} + \beta_{m}} \prod_{l=1}^{m-1} \left( 1 - \frac{\delta_{l}}{\delta_{l} + \beta_{l}} \right) \\
		& \times \frac{\gamma_{mk}}{\gamma_{mk} + \alpha_{mk}} \prod_{l=1}^{k-1} \left( 1 - \frac{\gamma_{ml}}{\gamma_{ml} + \alpha_{ml}} \right) \\
		& \times \op{T} \left( \vec{\mu}_{mk}, \frac{\kappa_{mk}}{1 + \kappa_{mk}} \mat{\Psi}_{m}, \nu_{m} + 1 \right).
	\end{aligned}
\end{align*}

\section{Empirical Evaluation}
\label{sec:results}
We evaluate the presented models on a range of tasks. Our goals are (1) to highlight the advantages of \gls{ILR} and \gls{HILR}, such as dealing with out-of-distribution predictions, recovering an input-dependent noise function, hierarchical gating, sharing parameters, and the ability to perform Bayesian sequential updates, (2) to benchmark the models on high dimensional datasets from real robots, and (3) to deploy the models in a real-world scenario to further empirically demonstrate its validity. A reference implementation can be found under \url{https://github.com/hanyas/mimo}.

\textbf{Out-of-distribution Uncertainty.} In Figure~\ref{fig:sine_cmb} (left), we apply \gls{ILR} on a synthetic dataset with two large gaps. We observe how the predictive uncertainty strongly reflects the lack of training data in these regions and how the mean prediction falls back to the prior values. This example highlights the model's reasonable uncertainty quantification. The out-of-distribution behavior of \gls{ILR} is strongly influenced by the discrete gating and a query's activation probability that jointly define the overall membership weights.

\begin{figure*}[t]
	\begin{minipage}[t]{0.45\textwidth}
		\includegraphics{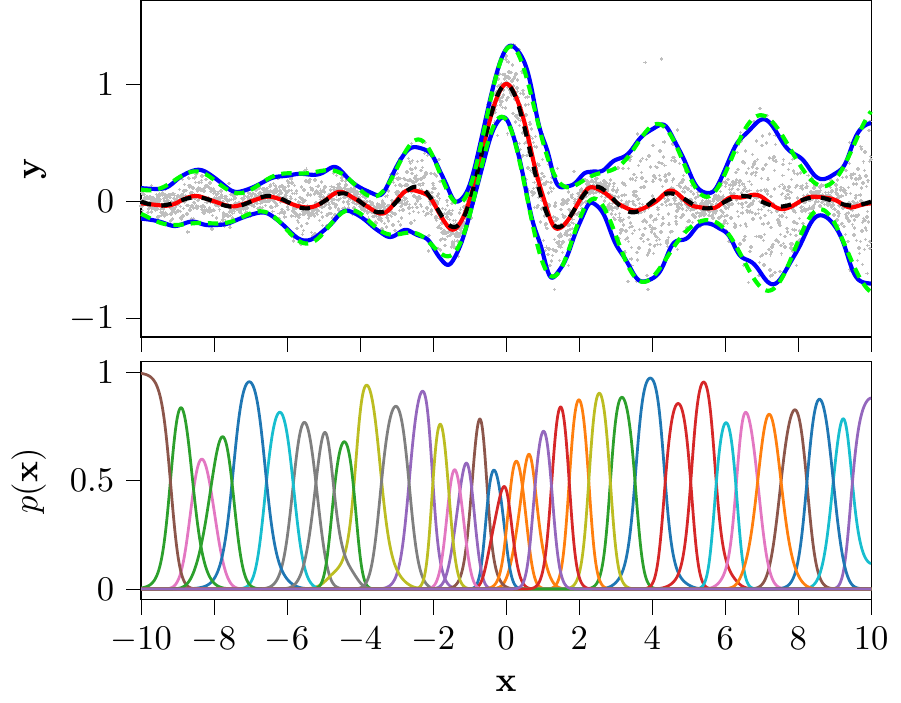}
	\end{minipage}\hspace{0.85cm}
	\begin{minipage}[t]{0.45\textwidth}
		\input{figures/triangle.tex}
	\end{minipage}
	\vspace{-0.75cm}%
	\caption{Left, a challenging heteroscedastic example of a sinc function heavily overlayed with input-dependent noise. The figure shows the mean prediction (red) on the training data (dots) and the true mean function (dashed black) corrupted by noise (dashed green). The blue dashed lines represent the recovered noise process. Right, hierarchical local regression with \acs{HILR} using parameter sharing in shift-invariant functions. The top figure shows the \emph{mode-prediction} (red) along with two standard deviations of predictive uncertainty (shaded blue). The bottom plot highlights the multi-modal activation, leading to shared slope information over non-adjacent regions.}
	\label{fig:sinc_triangle}
	\vspace{-0.1cm}
\end{figure*}

\textbf{Heteroscedastic Noise.} We test on two different problems with input-dependent noise, the \gls{CMB} \cite{bennett2003first}, Figure~\ref{fig:sine_cmb} (right), and a synthetic dataset from a stochastic sinc function $y(x)= \op{sinc}(x) + \epsilon$, where the noise $\epsilon$ is distributed according to zero-mean normal with a standard deviation $\sigma_{\epsilon}(x) = 0.05 + 0.2 (1 + \sin (2x)) /(1 + e^{-0.2 x})$. In Figure~\ref{fig:sinc_triangle} (left), we see that \gls{ILR} can approximate the nonlinear functions well. In particular, the heteroscedastic noise function is recovered in great detail.

\begin{figure*}[!]
	\begin{minipage}[t]{0.33\textwidth}
		\input{figures/step.tex}
	\end{minipage}\hfill
	\begin{minipage}[t]{0.3\textwidth}
		\input{figures/step_poly_features.tex}
	\end{minipage}\hfill
	\begin{minipage}[t]{0.3\textwidth}
		\input{figures/inverse_experts.tex}
	\end{minipage}
	\vspace{-0.65cm}
	\caption{Left and middle, learning discontinuous functions with \acs{ILR}. The top figures show the \textit{mode-prediction} (red) and two standard deviations confidence (shaded blue). The left example is a simple step function that can be captured with linear features, while in the middle, we use a polynomial transformation of the input for more flexibility. Right, we tackle inverse mapping problems with \acs{ILR}. This example includes scattered data that maps the input $\vec{x}$ to multiple output values $\vec{y}$. A discriminative modeling approach fails in these scenarios, as it tries to capture the ambiguous mean of the function $f: \vec{x} \rightarrow \vec{y}$. By approximating the joint density over both input and output and using \textit{mode-prediction}, \acs{ILR} can reconstruct these non-unique relations via local linear approximations. The bottom plots show the activation over the input space.}
	\label{fig:step_inverse}
	\vspace{-0.1cm}
\end{figure*}
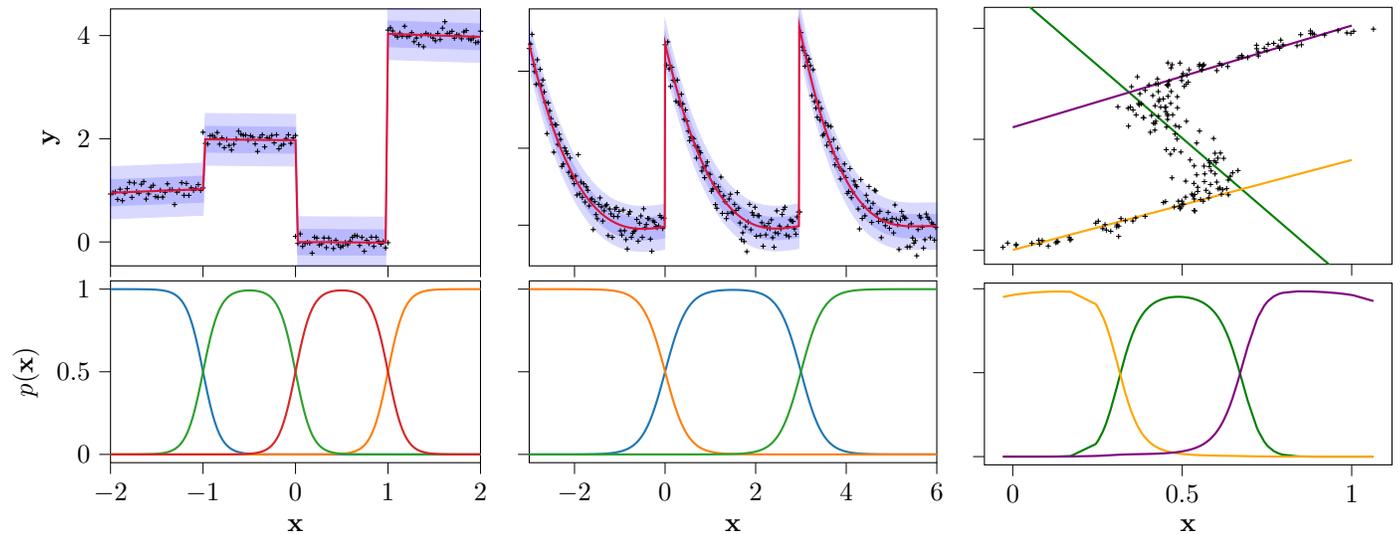

\textbf{Hierarchical Parameter Sharing.} In Figure~\ref{fig:sinc_triangle} (right), we test \gls{HILR}'s ability to share slope parameters via multi-modal activations. We consider a dataset stemming from a periodic triangle signal overlayed with additive noise. \gls{HILR} decides to activate two upper- and two lower-level regions to match the data structure, despite having more degrees of freedom at each level.

\textbf{Discontinuous (polynomial) Functions.} In Figure~\ref{fig:step_inverse} (left), a combination of step functions is fitted using mode-prediction, as described in Section~\ref{sec:ilr_predictive}. By using polynomial input features, more expressive local models can be realized. Figure~\ref{fig:step_inverse} (middle) depicts an example of cubic regressors, which are still linear in the parameters, fitted to data sampled from noisy cubic polynomials.

\textbf{Inverse Mapping.} One crucial advantage of generative over discriminative modeling is the ability to deal with non-unique inverse mapping problems. Such scenarios arise when the same input can be mapped to multiple output values. Joint modeling of the input-output data allows for flexible conditioning and alleviates the directional graph constraints. In Figure~\ref{fig:step_inverse} (right), we show a simple example of how \gls{ILR} can learn these mappings, adapted from \cite{murphy2012machine}.

\begin{figure*}[t]
	\centering
	\begin{minipage}{0.33\textwidth}
		\input{figures/chirp_0.tex}
	\end{minipage}\hspace{0.2cm}
	\begin{minipage}{0.3\textwidth}
		\input{figures/chirp_1.tex}
	\end{minipage}
	\begin{minipage}{0.3\textwidth}
		\input{figures/chirp_2.tex}
	\end{minipage}
	\vspace{-0.25cm}
	\caption{Bayesian sequential updates. Mean (red) and a two standard deviations interval (shaded blue) of the predictive distribution fitted to sequentially arriving data (three batches) from the chirp dataset (gray dots). For the second and third plots, the posterior fitted to the previous batches is used as a prior to perform a Bayesian sequential update. There is no catastrophic forgetting of previously learned knowledge, and in regions with no data, the prediction falls back to the prior.}
	\label{fig:bayes}
	\vspace{-0.25cm}
\end{figure*}
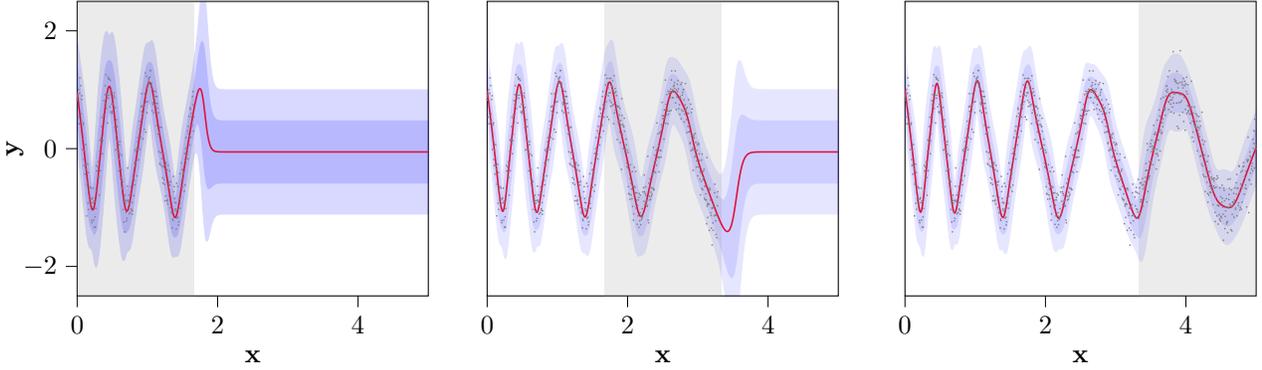

\textbf{Bayesian Sequential Updates.} In Figure~\ref{fig:bayes}, we demonstrate a sequential learning problem. Data from a chirp signal arrives in three batches so that the posterior of previous batches serves as the prior for the next learning phase. \Gls{ILR} successfully captures the data trend, and no significant catastrophic forgetting is observed. The approximation errors due to the mean-field posterior factorization assumption have little influence because the posterior updates are localized in the input domain.

\textbf{Robot Inverse Dynamics.} Next, we use \gls{ILR} and \gls{HILR} to learn the inverse dynamics of anthropomorphic manipulators. The dynamics of these manipulators are governed by the following equation
\begin{equation*}
	\vec{u} = \mat{M}(\vec{q}) \ddot{\vec{q}} + \mat{C}(\vec{q}, \dot{\vec{q}}) + \mat{G}(\vec{q}) + \epsilon(\vec{q}, \dot{\vec{q}}, \ddot{\vec{q}}),
\end{equation*}
where $(\vec{q}, \dot{\vec{q}}, \ddot{\vec{q}})$ are joint angles, velocities and accelerations, and $\vec{u}$ are torques. $\mat{M}(\vec{q})$ is the inertia matrix, $\mat{C}(\vec{q}, \dot{\vec{q}})$ are the Coriolis and centripetal forces, and $\mat{G}(\vec{q})$ is the gravity force. $\epsilon(\vec{q}, \dot{\vec{q}}, \ddot{\vec{q}})$ are general unmodelled nonlinearities such as sticktion/friction and hydraulic and tendon/cable dynamics, that motivate a data-driven approach to learn the mapping $(\vec{q}, \dot{\vec{q}}, \ddot{\vec{q}}) \rightarrow \vec{u}$. Later, we use the learned \gls{ILR} model for online inverse dynamics control.

\begin{table}[t]
	\centering
	\renewcommand{\arraystretch}{1}
	\caption{Accuracy on the SARCOS dataset}
	\sisetup{scientific-notation=fixed, table-align-uncertainty=true, separate-uncertainty=true}
	\setlength\tabcolsep{3.5pt}
	\footnotesize
	\begin{tabular}{c | S[fixed-exponent=-1] | S[fixed-exponent=-3] | S}
		\toprule[1.1pt]
		& {MSE}                 & {NMSE}                      & {Experts} \\ \midrule
		ILR             & \num{0.480 \pm 0.030} & \num{0.00340 \pm 0.0002}    & \num{1700} \\
		HILR            & \num{0.530 \pm 0.040} & \num{0.00390 \pm 0.0003}    & \num{1450} \\
		LGR$^{*}$       & \num{8.600 \pm 0.000} & \num{0.05000 \pm 0.0000}    & \num{7000} \\
		LWPR            & \num{2.600 \pm 0.030} & \num{0.01800 \pm 0.0002}    & \num{32000} \\
		GPR             & N/A & N/A & {-} \\
		SGPR            & \num{0.850 \pm 0.003} & \num{0.00600 \pm 0.000008}  & {-} \\
		rBCM            & \num{0.452 \pm 0.005} & \num{0.00260 \pm 0.000030}  & \num{315} \\
		gPoE            & \num{0.460 \pm 0.006} & \num{0.00300 \pm 0.000075}  & \num{315} \\
		\bottomrule[1.1pt]
	\end{tabular}
	\label{tab:sota_sarcos}
\end{table}

We use the \gls{MSE}, \gls{NMSE}, and the number of local experts as evaluation criteria. These measures cover the prediction accuracy as well as the complexity of the learned model. We compare to popular (probabilistic) methods such as \acf{LGR} \cite{meier2014incremental}, \acf{LWPR} \cite{vijayakumar2005incremental}, \acf{GPR} \cite{rasmussen2005gaussian} and \acf{SGPR} \cite{titsias2009variational}, and two scalable Gaussian process product of experts: the \gls{rBCM} \cite{deisenroth2015distributed} and the \gls{gPoE} \cite{cao2014generalized}.

We benchmark the prediction accuracy of all regression techniques on a high-dimensional dataset collected from a 7-DoF (degrees of freedom) anthropomorphic SARCOS arm \cite{vijayakumar2005incremental}. The dataset consists of $44484$ training points and 4449 test cases. Overall there are 21 input variables $(\vec{q}, \dot{\vec{q}}, \ddot{\vec{q}})$ mapping to 7 motor torques $\vec{u}$. We also benchmark on an inverse dynamics dataset from a 4-DoF Barrett-WAM manipulator, mapping from a 12-D to 4-D space. This dataset contains $25000$ training and $5000$ test pairs.

Table~\ref{tab:sota_sarcos} and Table~\ref{tab:sota_barrett} list the results for both datasets. We report the \gls{MSE}, \gls{NMSE}, and the number of active models over all joints, averaged over five seeds. We compute the mean and standard deviation for every cell in the table, except for \gls{LGR}$^{*}$, because of the unreasonable training times achieved while using the authors' code. When evaluating \gls{GPR} on the SARCOS dataset, we faced GPU-memory constraints $(\SI{32}{\giga\byte})$, and we have discarded this evaluation. For \gls{rBCM} and \gls{gPoE}, we assigned an expert to every 1000 data points, repeated for every output dimension.

The results show that \gls{ILR} and \gls{HILR} outperform the related local regression methods \gls{LWPR} and \gls{LGR}, both in terms of prediction accuracy and number of active experts. Nonetheless, \gls{GPR} is still the gold standard when the kernel size is within memory limits. Interestingly, the figures also indicate that \gls{ILR} and \gls{HILR} are competitive with \gls{SGPR} and the two product of experts \gls{rBCM} and \gls{gPoE}. Finally, the evidence reveals that \gls{HILR} tends to activate roughly $10-15\%$ fewer components than \gls{ILR}. This observation hints that \gls{HILR} may be taking advantage of shift-invariance patterns in the data and avoiding duplicate regression units. However, this hypothesis is hard to validate due to the data's high dimensionality.

\begin{table}[t]
	\centering
	\renewcommand{\arraystretch}{1}
	\caption{Accuracy on the Barrett-WAM dataset}
	\sisetup{scientific-notation=fixed, table-align-uncertainty=true, separate-uncertainty=true}
	\setlength\tabcolsep{3.5pt}
	\footnotesize
	\begin{tabular}{c | S[fixed-exponent=-1] | S[fixed-exponent=-3] | S}
		\toprule[1.1pt]
		& {MSE}                 & {NMSE}                   & {Experts} \\ \midrule
		ILR       & \num{0.290 \pm 0.050} & \num{0.0070 \pm 0.0005}   & \num{1350} \\
		HILR      & \num{0.310 \pm 0.065} & \num{0.0080 \pm 0.0006}   & \num{1110} \\
		LGR$^{*}$ & \num{0.770 \pm 0.000} & \num{0.0170 \pm 0.0000}   & \num{3270} \\
		LWPR      & \num{1.000 \pm 0.150} & \num{0.0370 \pm 0.0100}   & \num{2900} \\
		GPR       & \num{0.100 \pm 0.001} & \num{0.0023 \pm 0.00001}  & {-} \\
		SGPR      & \num{0.180 \pm 0.005} & \num{0.0063 \pm 0.00002}  & {-} \\
		rBCM      & \num{0.380 \pm 0.035} & \num{0.0190 \pm 0.00180}  & \num{100} \\
		gPoE      & \num{0.340 \pm 0.013} & \num{0.0160 \pm 0.00060}  & \num{100} \\
		\bottomrule[1.1pt]
	\end{tabular}
	\label{tab:sota_barrett}
\end{table}

\textbf{Real Inverse Dynamics Control.} Finally, we demonstrate the validity of the learned dynamics in a real-robot control scenario. In this evaluation, we tackle two experiments. In the first, we focus on the Bayesian sequential learning aspect, whereas the second highlights our model's ability to deal with massive amounts of data. Both experiments deploy an \gls{ILR} model on the Barrett-WAM to track an 8-shaped trajectory in the $xy$-plane of the end-effector.

\begin{figure}[t]
	\centering
	\input{figures/eight_seq_smse.tex}
	\caption{8-Shaped trajectory learning. Bayesian sequential updates on a dataset collected from a Barrett-WAM. We plot the \gls{NMSE} for five different seeds on accumulated data over the number of batches. The \gls{NMSE} consistently improves with new data, and no catastrophic forgetting is observed.}
	\label{fig:eight_bayesian}
\end{figure}
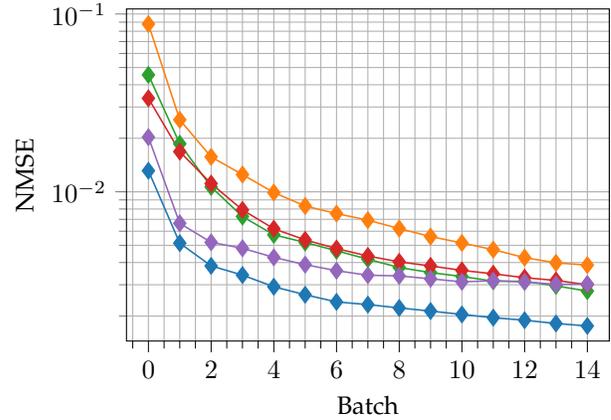

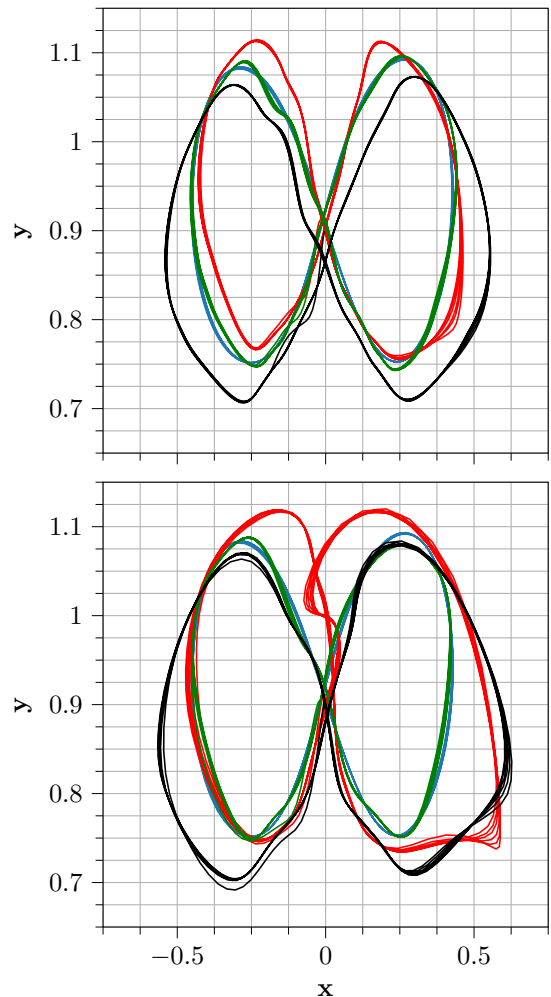
\begin{figure}[!]
	\begin{minipage}{0.475\textwidth}
		\centering
		\input{figures/eight_0275.tex}
	\end{minipage}\hfill%
	\begin{minipage}{0.475\textwidth}
		\centering
		\input{figures/eight_0425.tex}
	\end{minipage}\hfill%
	\vspace{-0.25cm}
	\caption{8-shaped trajectory tracking on the Barrett-WAM. We compare three controllers on two test trajectories (blue), a low-gain PD (black), a low-gain PD + feed-forward torques from an analytical model (red), and a low-gain PD + feed-forward torques from \acs{ILR} (green). The results indicate that \acs{ILR} delivers the best tracking performance.}
	\label{fig:eight_tracking}
	\vspace{-0.25cm}
\end{figure}

For the first experiment, we collect $30000$ training samples (roughly $1$ minute) from multiple trajectories with different velocity profiles and Bayesian sequential learning over $15$ batches for multiple seeds. Figure~\ref{fig:eight_bayesian} depicts the progression of the learning process. We then select the best model and perform model-based control to track held-out test trajectories with unseen velocity profiles. \Gls{ILR} predicts the feed-forward torques while supported by a low-gain PD-controller for safety considerations. We compare this controller to one with access to the analytical dynamics and the same low-gain PD-controller and to a \emph{model-free} PD-controller. Figure~\ref{fig:eight_tracking} depicts the tracking performance of the different controllers on two test trajectories.

In the second experiment, we learn a model covering a larger region of the state-action space. We generate a larger Barrett-WAM dataset consisting of $150000$ training examples (roughly 5 minutes). We test the model learned by \gls{ILR} on held-out test trajectories and compare it to the same controller constellation as in the first experiment. As benchmarking criteria, we evaluate the \gls{MSE} with respect to the desired trajectory and the mean torque contribution of the PD-controller to the overall control signal. The rationale is that a good inverse dynamics model will consistently produce a low \gls{MSE} without relying on the PD-controller's assistance. Table~\ref{tab:invdyn} shows the benchmarks on three test trajectories. The results indicate that \gls{ILR} significantly improves tracking performance with minimal contribution from the PD-controller. Moreover, we are able to consistently achieve a prediction frequency of $\SI{2000}{\Hz}$ during both tasks, although the Barrett-WAM robot requires only $\SI{500}{\Hz}$.

\begin{table}[!]
	\caption{Tracking error and torque contributed by the PD-controller during the Barrett-WAM real robot task.}
	\vspace{-0.25cm}
	\footnotesize
	\begin{center}
		\setlength\tabcolsep{6pt}
		\begin{tabular}{c | c | c | c | c }
			\toprule[1.1pt]
			&           & PD                    & Analytic+PD           & ILR+PD \\ \midrule
			\multirow{2}{*}{T1} & MSE       & $2.33 \times 10^{-2}$ & $2.16 \times 10^{-2}$ & $1.03 \times 10^{-3}$ \\
			& PD-Torque & $8.25 \times 10^{0}$  & $7.12 \times 10^{0}$  & $1.40 \times 10^{0}$ \\
			\midrule       
			\multirow{2}{*}{T2} & MSE       & $2.60 \times 10^{-2}$ & $2.55 \times 10^{-2}$ & $9.17 \times 10^{-4}$ \\
			& PD-Torque & $8.71 \times 10^{0}$  & $7.41 \times 10^{0}$  & $1.33 \times 10^{0}$ \\
			\midrule
			\multirow{2}{*}{T3} & MSE       & $2.94 \times 10^{-2}$ & $3.08 \times 10^{-2}$ & $8.96 \times 10^{-4}$ \\
			& PD-Torque & $9.38 \times 10^{0}$  & $8.06 \times 10^{0}$  & $1.38 \times 10^{0}$ \\
			\bottomrule[1.1pt]
		\end{tabular}
	\end{center}
	\label{tab:invdyn}
	\vspace{-0.25cm}
\end{table}

\section{Conclusion}
\label{sec:conclusion}
In this paper, we presented two probabilistic hierarchical local regression models, \gls{ILR} and \gls{HILR}, and derived an efficient variational inference technique for data-driven learning. These representations are based on infinite mixtures from Bayesian nonparametrics. We situate our contributions as the next iteration in a large family of local linear regression techniques such as \gls{RFWR}, \gls{LWPR}, and \gls{LGR}. We have shown that placing Dirichlet process priors on Bayesian mixtures of local regression units can regularize model complexity with minor loss in performance and without relying on heuristics. Moreover, we have highlighted the advantage of the two-level architecture of \gls{HILR} in compressing the learned models and reducing the number of active experts. Empirical evaluation indicates that the models outperform \gls{LWPR} and \gls{LGR}, and are competitive with sparse \gls{GPR} and products of experts. Finally, we have empirically confirmed the practicality of this approach for online inverse dynamics control on a Barrett-WAM robot.

Nonetheless, these presented concepts still suffer from multiple drawbacks. The mean-field assumption is a source of significant errors in posterior inference. Collapsed formulations of Dirichlet process priors promise better approximations \cite{kurihara2007collapsed}. In addition, Bayesian mixture models are influenced by a large number of hyperparameters, which cannot be directly optimized via empirical Bayes \cite{maritz1989empirical}, leading to lower predictive performance when compared to optimized \gls{GPR}. Nonetheless, the \gls{ELBO} offers an objective for hyperparameter optimization. Naive gradient-based techniques have proven too brittle due to their reliance on Euclidean distance metrics. A natural-gradient approach appears to be a suitable alternative in the future.

Further development of hierarchical local regression may focus on treating \gls{ILR} and \gls{HILR} as layers in a multi-layered representation. This extension would allow the models to benefit from intermediate nonlinear projections into high dimensional spaces that have proven powerful in deep neural networks. Another practical consideration is to incorporate physical inductive biases such as inverse dynamics \cite{nguyen2010using} to facilitate learning meaningful quantities.

\section*{Acknowledgments}
This work has received funding from the European Unions Horizon 2020 research and innovation program under grant agreement No.~640554 (SKILLS4ROBOTS).

\appendices
\section{}
In the following, we present more detailed derivations of the E-step and M-step of \acrshort{ILR} and \acrshort{HILR}.
\subsection*{E-Step of Infinite Linear Regression}
\label{app:ilr_estep}
\begin{align*}
	\log q(\mat{Z}) & = 
	\begin{aligned}[t]
		& \mathbb{E}_{q(\vec{s})} \left[ \log p(\mat{Z} | \vec{\pi}(\vec{s})) \right] \\
		& + \mathbb{E}_{q(\vec{\mu}, \mat{\Lambda})} \left[ \log p(\mat{X} | \mat{Z}) \right] \\
		& + \mathbb{E}_{q(\mat{A}, \vec{c}, \mat{V})} \left[ \log p(\mat{Y} | \mat{Z}, \mat{X}) \right] + \const
	\end{aligned} \\
	& \hspace{-1.0cm} =
	\begin{aligned}[t]
		& \const + \sum_{n=1}^{N} \Big[ \mathbb{E}_{q(\vec{s})} \left[ \log \Cat(\vec{z}_n | \vec{\pi}(\vec{s})) \right] \\
		& \! + \! \sum_{k=1}^{K} z_{nk} \mathbb{E}_{q(\vec{\mu}, \mat{\Lambda})} \left [ \log \N(\mat{x}_{n} | \vec{\mu}_{k}, \mat{\Lambda}_{k} ) \right] \\
		& \! + \! \sum_{k=1}^{K} z_{nk} \mathbb{E}_{q(\mat{A}, \vec{c}, \mat{V})} \left[ \log \N(\vec{y}_{n} | \mat{A}_{k} \vec{x}_{n} + \vec{c}_{k}, \mat{V}_{k})  \right] \Big]
	\end{aligned} \\
	& \hspace{-1.0cm} = \sum_{n=1}^{N} \sum_{k=1}^{K} z_{nk} \log r_{nk},
\end{align*}
where
\begin{align*}
	\mathbb{E}_{q(\vec{s})} \left[ \log \Cat(\vec{z}_n | \vec{\pi}(\vec{s})) \right] & \! = \!
	\begin{aligned}[t]
		& \sum_{k=1}^{K} z_{nk} \mathbb{E}_{q(s_{k})} \left[ \log s_{k}  \right] \\
		& + \sum_{k=1}^{K} z_{nk} \mathbb{E}_{q(s_{k})} \left[ \sum_{l=1}^{k-1} \log (1 - s_{l}) \right].
	\end{aligned}
\end{align*}
The individual expectations of the \emph{log-sticks} under the beta posteriors are given by
\begin{align*}
	\mathbb{E}_{q(s_{k})} \left[ \log s_{k} \right]       & = \Psi \left( \gamma_{k} \right) - \Psi \left( \gamma_{k} + \alpha_{k} \right), \\
	\mathbb{E}_{q(s_{k})} \left[ \log (1 - s_{k}) \right] & = \Psi \left( \alpha_{k} \right) - \Psi \left( \gamma_{k} + \alpha_{k} \right),
\end{align*}
where $\Psi$ is the Digamma function.

\subsection*{M-Step of Infinite Linear Regression}
\label{app:ilr_mstep}
\begin{align*}
	\log q(\vec{s}) & = \mathbb{E}_{q(\mat{Z})} \left[ \log  p(\mat{Z} | \vec{\pi}(\vec{s})) \right] + \log p(\vec{s}) + \const \\
	& = 
	\begin{aligned}[t]
		& \sum_{n=1}^{N} \sum_{k=1}^{K} r_{nk} \log \left[ s_{k} \prod_{l=1}^{k-1} (1 - s_{l}) \right] \\
		& + \sum_{k=1}^{K-1} \log \Beta (s_{k} | 1, \alpha_{0}) + \const,
	\end{aligned}
\end{align*}
\begin{align*}
	\log q(\vec{\mu}, \mat{\Lambda}) & = \mathbb{E}_{q(\mat{Z})} \left[ \log p(\mat{X} | \mat{Z}) \right] + \log p(\vec{\mu}, \mat{\Lambda}) + \const \\
	& = 
	\begin{aligned}[t]
		& \sum_{n=1}^{N} \sum_{k=1}^{K} r_{nk} \log \N(\mat{x}_{n} | \vec{\mu}_{k}, \mat{\Lambda}_{k}) \\
		& + \sum_{k=1}^{K} \log \N (\mat{\mu}_{k} | \mat{m}_{0}, \kappa_{0} \mat{\Lambda}_{k}) \\
		& + \sum_{k=1}^{K} \log \W(\mat{\Lambda}_{k} | \mat{\Psi}_{0}, \nu_{0})  + \const,
	\end{aligned} \\
	\log q(\mat{A}, \vec{c}, \mat{V}) & =
	\begin{aligned}[t]
		& \mathbb{E}_{q(\mat{Z})} \left[ \log p(\mat{Y} | \mat{Z}, \mat{X}) \right] \\
		& + \log p(\mat{A}, \vec{c}, \mat{V}) + \const
	\end{aligned} \\
	&  =
	\begin{aligned}[t]
		& \sum_{n=1}^{N} \sum_{k=1}^{K} r_{nk} \log \N(\vec{y}_{n} | \mat{A}_{k} \vec{x}_{n} + \vec{c}_{k}, \mat{V}_{k}) \\ 
		& + \sum_{k=1}^{K} \log \MN(\mat{A}_{k} | \mat{M}_{0}, \mat{K}_{0}, \mat{V}_{k}) \\
		& + \sum_{k=1}^{K} \log \N(\vec{c}_{k} | \vec{\theta}_{0}, \rho_{0} \mat{V}_{k}) \\
		& + \sum_{k=1}^{K} \log \W(\mat{V}_{k} | \mat{\Phi}_{0}, \eta_{0}) + \const,
	\end{aligned}
\end{align*}
where $r_{nk} = \mathbb{E}_{q(\mat{Z})} \left[ z_{nk} \right]$ are the expected responsibilities computed in the E-step. Further details on performing these updates via general exponential family recipes are provided in the upcoming sections.

\subsection*{M-Step of Hierarchical Infinite Linear Regression}
\label{app:hilr_mstep}
\begin{align*}
	& \log q(\vec{t})
	\begin{aligned}[t]
		& = \mathbb{E}_{q(\mat{H})} \left[ \log  p(\mat{H} | \vec{\omega}(\vec{t})) \right] + \log p(\vec{t}) + \const \\
		& = 
		\begin{aligned}[t]
			& \sum_{n=1}^{N} \sum_{m=1}^{M} \hat{g} \log \left[ t_{m} \prod_{l=1}^{m-1} (1 - t_{l}) \right] \\
			& + \sum_{m=1}^{M} \log \Beta (t_{m} | 1, \beta_{0}) + \const,
		\end{aligned}
	\end{aligned} \\
	& \log q(\vec{s})
	\begin{aligned}[t]
		& = \mathbb{E}_{q(\mat{H}, \mat{Z})} \left[ \log p(\mat{Z} | \mat{H}) \right] + \log p(\vec{s}) + \const \\
		& = 
		\begin{aligned}[t]
			& \sum_{n=1}^{N} \sum_{m=1}^{M} \sum_{k=1}^{K} \hat{g} \, \hat{r} \log \left[ s_{mk} \prod_{l=1}^{k-1} (1 - s_{ml}) \right] \\
			& + \sum_{m=1}^{M} \sum_{k=1}^{K} \log \Beta (s_{mk} | 1, \alpha_{0}) + \const,
		\end{aligned}
	\end{aligned} \\
	& \log q(\vec{\mu}) 
	\begin{aligned}[t]
		& =
		\begin{aligned}[t]
			& \mathbb{E}_{q(\mat{H}, \mat{Z}, \vec{\tau}, \mat{\Lambda})} \left[ \log p(\mat{X} | \mat{H}, \mat{Z}) \right] \\
			& + \mathbb{E}_{q(\vec{\tau}, \mat{\Lambda})} \left[ \log p(\vec{\mu} | \vec{\tau}, \mat{\Lambda}) \right] + \const
		\end{aligned} \\
		& =
		\begin{aligned}[t]
			& \sum_{n=1}^{N} \sum_{m=1}^{M} \sum_{k=1}^{K} \hat{g} \, \hat{r} \, \mathbb{E}_{q(\mat{\Lambda})} \left[ \log \N(\vec{x}_{n} | \vec{\mu}_{mk}, \mat{\Lambda}_{m}) \right] \\
			& + \sum_{m=1}^{M} \sum_{k=1}^{K} \mathbb{E}_{q(\vec{\tau}, \mat{\Lambda})} \left[ \log \N(\vec{\mu}_{mk} | \vec{\tau}_{m}, \kappa_{0} \mat{\Lambda}_{m}) \right] \\
			& + \const,
		\end{aligned}
	\end{aligned} \\
	& \log q(\vec{\tau}, \mat{\Lambda}) 
	\begin{aligned}[t]
		& = 
		\begin{aligned}[t]
			& \mathbb{E}_{q(\mat{H}, \mat{Z}, \vec{\mu})} \left[ \log p(\mat{X} | \mat{H}, \mat{Z}) \right] + \log p(\vec{\tau}, \mat{\Lambda}) \\
			& + \mathbb{E}_{q(\vec{\mu})} \left[ \log p(\vec{\mu} | \vec{\tau}, \mat{\Lambda}) \right] + \const
		\end{aligned} \\
		& = 
		\begin{aligned}[t]
			& \sum_{n=1}^{N} \sum_{m=1}^{M} \sum_{k=1}^{K} \hat{g} \, \hat{r} \, \mathbb{E}_{q(\vec{\mu})} \left[ \log \N(\vec{x}_{n} | \vec{\mu}_{mk}, \mat{\Lambda}_{m}) \right] \\
			& + \sum_{m=1}^{M} \log \N(\vec{\tau}_{m} | \vec{m}_{0}, \lambda_{0} \mat{\Lambda}_{m}) \\ 
			& + \sum_{m=1}^{M} \log \W(\mat{\Lambda}_{m} | \mat{\Psi}_{0}, \nu_{0}) \\
			& + \sum_{m=1}^{M} \sum_{k=1}^{K} \mathbb{E}_{q(\vec{\mu})} \left[ \log \N(\vec{\mu}_{mk} | \vec{\tau}_{m}, \kappa_{0} \mat{\Lambda}_{m}) \right] \\
			& + \const,
		\end{aligned}
	\end{aligned} \\
	& \log q(\vec{c}) 
	\begin{aligned}[t]
		& =
		\begin{aligned}[t]
			& \mathbb{E}_{q(\mat{H}, \mat{Z}, \mat{A}, \mat{V})} \left[ \log p(\mat{Y} | \mat{H}, \mat{Z}, \mat{X}) \right] \\
			& + \mathbb{E}_{q(\mat{V})} \left[ \log p(\vec{c} | \mat{V}) \right] + \const
		\end{aligned} \\
		& = 
		\begin{aligned}[t]
			& \sum_{n=1}^{N} \sum_{m=1}^{M} \sum_{k=1}^{K} \hat{g} \, \hat{r} \, \\
			& \times \mathbb{E}_{q(\mat{A}, \mat{V})} \left[ \log \N(\vec{y}_{n} | \mat{A}_{m} \vec{x}_{n} + \vec{c}_{mk}, \mat{V}_{m}) \right] \\
			& + \sum_{m=1}^{M} \sum_{k=1}^{K} \mathbb{E}_{q(\mat{V})} \left[ \MN(\vec{c}_{mk} | \vec{\theta}_{0}, \rho_{0} \mat{V}_{m}) \right] \\
			& + \const,
		\end{aligned}
	\end{aligned} \\
	& \log q(\mat{A}, \mat{V})
	\begin{aligned}[t]
		& =     
		\begin{aligned}[t]
			& \mathbb{E}_{q(\mat{H}, \mat{Z}, \vec{c})} \left[ \log p(\mat{Y} | \mat{H}, \mat{Z}, \mat{X}) \right] + \log p(\mat{A}, \mat{V}) \\
			& + \mathbb{E}_{q(\vec{c})} \left[ \log p(\vec{c} | \mat{V}) \right] + \const
		\end{aligned} \\
		& = 
		\begin{aligned}[t]
			& \sum_{n=1}^{N} \sum_{m=1}^{M} \sum_{k=1}^{K} \hat{g} \, \hat{r} \, \\
			& \times \mathbb{E}_{q(\vec{c})} \left[ \log \N(\vec{y}_{n} | \mat{A}_{m} \vec{x}_{n} + \vec{c}_{mk}, \mat{V}_{m}) \right] \\
			& + \sum_{m=1}^{M} \sum_{k=1}^{K} \mathbb{E}_{q(\vec{c})} \left[ \MN(\vec{c}_{mk} | \vec{\theta}_{0}, \rho_{0} \mat{V}_{m}) \right] \\
			& + \const,
		\end{aligned}
	\end{aligned}
\end{align*}
where the quantities $\hat{g} \,=\, g_{nm} \,=\, \mathbb{E}_{q(\mat{H})} \left[ h_{nm} \right]$ and $\hat{r} = r_{nmk} =  \mathbb{E}_{q(\mat{Z})} \left[ z_{nmk} \right]$. After computing the necessary expectations, these updates correspond to the posterior update recipes described in later appendix sections.

\section{}
\label{app:exp}
Our work mainly considers random variables with probability density functions belonging to the exponential family. The unified minimal parameterization of this class of distributions lends itself to convenient and efficient posterior computation when paired with conjugate priors.

We assume the natural form for a probability density of a random variable $\vec{x}$
\begin{equation*}
	f(\vec{x} | \vec{\eta}) = h(\vec{x}) \exp \left[\vec{\eta} \cdot \vec{t}(\vec{x}) - a(\vec{\eta}) \right],
\end{equation*}
where $h(\vec{x})$ is the base measure, $\vec{\eta}$ are the natural parameters, $\vec{t}(\vec{x})$ are the sufficient statistics and $a(\vec{\eta})$ is the log-partition function, or log-normalizer. Following the same notation, a conjugate prior $g(\vec{\eta} | \vec{\lambda})$ to the likelihood $f(\vec{x} | \vec{\eta})$ has the form
\begin{equation*}
	g(\vec{\eta} | \vec{\lambda}) = h(\vec{\eta}) \exp \left[\vec{\lambda} \cdot \vec{t}(\vec{\eta}) - a(\vec{\lambda}) \right],
\end{equation*}
with prior sufficient statistics $\vec{t}(\vec{\eta}) = [\vec{\eta}, \, - a(\vec{\eta})]^{\top}$ and hyperparameters $\vec{\lambda} = [\vec{\alpha}, \, \vec{\beta}]^{\top}$. By applying Bayes' rule, we can directly infer the posterior $q(\vec{\eta} | \vec{x})$
\begin{align*}
	q(\vec{\eta} | \vec{x}) & \propto f(\vec{x} | \vec{\eta}) g(\vec{\eta} | \vec{\lambda})                                            \\
	& \propto \exp \left[\vec{\rho}(\vec{x}, \vec{\lambda}) \cdot \vec{t}(\vec{\eta}) - a(\vec{\rho}) \right],
\end{align*}
where the posterior natural parameters $\vec{\rho}(\vec{x}, \vec{\lambda})$ are a function of the likelihood sufficient statistics $\vec{t}(\vec{x})$ and prior hyperparameters $[\vec{\alpha}, \, \vec{\beta}]$
\begin{equation*}
	\vec{\rho}(\vec{x}, \vec{\lambda}) = \left[\vec{\alpha} + \vec{t}(\vec{x}), ~ \vec{\beta} + \vec{1} \right]^{\top}.
\end{equation*}
The structure of the resulting posterior reveals a simple recipe for data-driven inference. By moving into the natural space, the posterior parameters are computed by combining the prior hyperparameters with the likelihood's sufficient statistics and log-partition function. By definition, every exponential family distribution has a minimal natural parameterization that leads to a unique decomposition of these quantities \cite{wainwright2008graphical}.

\section{}
\label{app:posteriors}
We present an outline of all M-step updates. We use an adapted form of the exponential natural parameterization, as it offers a clear methodology on how to derive and implement such updates for all relevant distributions.

\subsection*{Gaussian with Normal-Wishart Prior}
A weighted Gaussian likelihood over a random variable $\vec{x} \in \mathds{R}^{d}$ has the following precision-based parameterization
\begin{align*}
	p(\mat{X} | \vec{\mu}, \mat{\Lambda}) & = \prod_{n=1}^{N} \N(\vec{x}_{n} | \vec{\mu}, \mat{\Lambda})^{w_{n}} \\
	& \propto
	\exp \left\{
	\begin{bmatrix}
		\mat{\Lambda} \vec{\mu}                  \\[0.5em]
		\vec{\mu}^{\top} \mat{\Lambda} \vec{\mu} \\[0.5em]
		\mat{\Lambda}                            \\[0.5em]
		\log |\mat{\Lambda}|
	\end{bmatrix}
	\! \cdot \!
	\begin{bmatrix}
		\sum_{n=1}^{N} w_{n} \vec{x}_{n}                                 \\[0.5em]
		-\frac{1}{2} \sum_{n=1}^{N} w_{n}                                \\[0.5em]
		-\frac{1}{2} \sum_{n=1}^{N} w_{n} \vec{x}_{n} \vec{x}_{n}^{\top} \\[0.5em]
		\frac{1}{2} \sum_{n=1}^{N} w_{n}
	\end{bmatrix}
	\right\},
\end{align*}
where $w_{n}$ are the importance weights. The conjugate prior $p(\vec{\mu}, \mat{\Lambda})$ is a normal-Wishart distribution with zero mean
\begin{align*}
	p(\vec{\mu}, \mat{\Lambda}) & = \N(\vec{\mu} | \vec{0}, \kappa_{0} \mat{\Lambda}) \W(\mat{\Lambda} | \mat{\Psi}_{0}, \nu_{0}) \\
	& \propto
	\exp \left\{
	\begin{bmatrix}
		\vec{0}                          \\[0.5em]
		-\frac{1}{2} \kappa_{0}          \\[0.5em]
		-\frac{1}{2} \mat{\Psi}_{0}^{-1} \\[0.5em]
		\frac{1}{2} (\nu_{0} - d)
	\end{bmatrix}
	\! \cdot \!
	\begin{bmatrix}
		\mat{\Lambda} \vec{\mu}                  \\[0.5em]
		\vec{\mu}^{\top} \mat{\Lambda} \vec{\mu} \\[0.5em]
		\mat{\Lambda}                            \\[0.5em]
		\log |\mat{\Lambda}|
	\end{bmatrix}
	\right\}.
\end{align*}
The resulting posterior $q(\vec{\mu}, \mat{\Lambda})$ is also a normal-Wishart
\begin{align*}
	q(\vec{\mu}, \mat{\Lambda}) & \! = \! \N(\vec{\mu} | \vec{m}, \kappa \mat{\Lambda}) \W(\mat{\Lambda} | \mat{\Psi}, \nu) \\
	\!                          & \! \propto
	\exp \left\{ \!
	\begin{bmatrix}
		\sum_{n=1}^{N} w_{n} \vec{x}_{n}                                                         \\[0.5em]
		-\frac{1}{2} (\kappa_{0} + \sum_{n=1}^{N} w_{n})                                         \\[0.5em]
		-\frac{1}{2} (\mat{\Psi}_{0}^{-1} + \sum_{n=1}^{N} w_{n} \vec{x}_{n} \vec{x}_{n}^{\top}) \\[0.5em]
		\frac{1}{2} (\nu_{0} - d + \sum_{n=1}^{N} w_{n})
	\end{bmatrix}
	\! \cdot \!
	\begin{bmatrix}
		\mat{\Lambda} \vec{\mu}                  \\[0.5em]
		\vec{\mu}^{\top} \mat{\Lambda} \vec{\mu} \\[0.5em]
		\mat{\Lambda}                            \\[0.5em]
		\log |\mat{\Lambda}|
	\end{bmatrix}
	\! \right\}.
\end{align*}
The resulting posterior parameters are
\begin{gather*}
	\kappa = \kappa_{0} + \sum_{n=1}^{N} w_{n}, \\
	\vec{m} = 1 / \kappa \sum_{n=1}^{N} w_{n} \vec{x}_{n}, \\
	\nu = \nu_{0} + \sum_{n=1}^{N} w_{n}, \\
	\mat{\Psi} = (\mat{\Psi}_{0}^{-1} + \sum_{n=1}^{N} w_{n} \vec{x}_{n} \vec{x}_{n}^{\top} - \kappa \, \vec{m} \, \vec{m}^{\top})^{-1}.
\end{gather*}

\subsection*{Linear-Gaussian with Matrix-Normal-Wishart Prior}
A weighted linear-Gaussian likelihood takes a random input variable $\vec{x} \in \mathds{R}^{d}$ and returns a response random variable $\vec{y} \in \mathds{R}^{m}$ according to a linear mapping $\mat{A}: \mathds{R}^{d} \to \mathds{R}^{m}$
\begin{align*}
	p(\mat{Y} | \mat{X}, \mat{A}, \mat{V}) & = \prod_{n=1}^{N} \N(\vec{y}_{n} | \vec{x}_{n}, \mat{A}, \mat{V})^{w_{n}} \\
	& \propto
	\exp \left\{
	\begin{bmatrix}
		\mat{V} \mat{A}                \\[0.5em]
		\mat{A}^{\top} \mat{V} \mat{A} \\[0.5em]
		\mat{V}                        \\[0.5em]
		\log |\mat{V}|
	\end{bmatrix}
	\cdot
	\begin{bmatrix}
		\mat{Y} \mat{W} \mat{X}^{\top}              \\[0.5em]
		-\frac{1}{2} \mat{X} \mat{W} \mat{X}^{\top} \\[0.5em]
		-\frac{1}{2} \mat{Y} \mat{W} \mat{Y}^{\top} \\[0.5em]
		\frac{1}{2} \sum_{n=1}^{N} w_{n}
	\end{bmatrix}
	\right\},
\end{align*}
where $w_{n}$ are the weights and $\mat{W} = \mathrm{diag}(w_{n})$ is the diagonal weight matrix. The data matrices $\mat{X}$ and $\mat{Y}$ are of size $d \times N$ and $m \times N$, respectively. The conjugate prior $p(\mat{A}, \mat{V})$ is a matrix-normal-Wishart with zero mean
\begin{align*}
	p(\mat{A}, \mat{V}) & = \N(\mat{A} | \mat{0}, \mat{V}, \mat{K}_{0}) \W(\mat{V} | \mat{\Psi}_{0}, \nu_{0}) \\
	& \propto
	\exp \left\{
	\begin{bmatrix}
		\mat{0}                          \\[0.5em]
		-\frac{1}{2} \mat{K}_{0}         \\[0.5em]
		-\frac{1}{2} \mat{\Psi}_{0}^{-1} \\[0.5em]
		\frac{1}{2} (\nu_{0} - m - 1 + d)
	\end{bmatrix}
	\cdot
	\begin{bmatrix}
		\mat{V} \mat{A}                \\[0.5em]
		\mat{A}^{\top} \mat{V} \mat{A} \\[0.5em]
		\mat{V}                        \\[0.5em]
		\log |\mat{V}|
	\end{bmatrix}
	\right\}.
\end{align*}
The posterior $q(\vec{\mu}, \mat{\Lambda})$ is matrix-normal-Wishart
\begin{align*}
	q(\mat{A}, \mat{V}) & = \N(\mat{A} | \mat{M}, \mat{V}, \mat{K}) \W(\mat{V} | \mat{\Psi}, \nu) \\
	& \hspace{-2.5em} \! \propto \!
	\exp \left\{ \!
	\begin{bmatrix}
		\mat{Y} \mat{W} \mat{X}^{\top}                                      \\[0.5em]
		-\frac{1}{2} (\mat{K}_{0} + \mat{X} \mat{W} \mat{X}^{\top})         \\[0.5em]
		-\frac{1}{2} (\mat{\Psi}_{0}^{-1} + \mat{Y} \mat{W} \mat{Y}^{\top}) \\[0.5em]
		\frac{1}{2} (\nu_{0} - m - 1 + d + \sum_{n=1}^{N} w_{n})
	\end{bmatrix}
	\! \cdot \!
	\begin{bmatrix}
		\mat{V} \mat{A}                \\[0.5em]
		\mat{A}^{\top} \mat{V} \mat{A} \\[0.5em]
		\mat{V}                        \\[0.5em]
		\log |\mat{V}|
	\end{bmatrix}
	\! \right\}.
\end{align*}
The standard-form posterior parameters are
\begin{gather*}
	\mat{K} = \mat{K}_{0} + \mat{X} \mat{W} \mat{X}^{\top}, \\
	\mat{M} = \mat{Y} \mat{W} \mat{X}^{\top} \mat{K}^{-1}, \\
	\nu = \nu_{0} + \sum_{n=1}^{N} w_{n}, \\
	\mat{\Psi} = (\mat{\Psi}_{0}^{-1} + \mat{Y} \mat{W} \mat{Y}^{\top} - \mat{M} \, \mat{K} \, \mat{M}^{\top} )^{-1}.
\end{gather*}

\subsection*{Infinite Categorical with Stick-Breaking Prior}
We assume one-hot random variable $\vec{z}$ of infinite size with a categorical likelihood
\begin{align*}
	p(\mat{Z} | \vec{\pi}(\vec{s})) & = \prod_{n=1}^{N} \Cat(\mat{z}_{n} | \vec{\pi}(\vec{s}))^{w_{nk}} \\
	& = \prod_{n=1}^{N} \prod_{k=1}^{\infty} \left[ \left( s_{k} \prod_{l=1}^{k-1} \left( 1 - s_{l} \right) \right)^{z_{nk}} \right]^{w_{nk}},
\end{align*}
where $w_{n}$ are the weights. The prior is an infinitely factorized beta distribution over the stick lengths $s_{k}$
\begin{align*}
	p(\vec{s}) & = \prod_{k=1}^{\infty} \Beta(s_k | \gamma_{0}, \alpha_{0}) \\
	& \propto \prod_{k=1}^{\infty} s_{k}^{\gamma_{0} - 1} \left( 1 - s_k \right)^{\alpha_{0} - 1},
\end{align*}
and the posterior is also a factorized beta density truncated up to a level $K$ \cite{blei2006variational}
\begin{equation*}
	q(\vec{s}) = \prod_{k=1}^{K} \Beta(s_k | \gamma_{k}, \alpha_{k}),
\end{equation*}
where
\begin{align*}
	\gamma_{k} & =  \gamma_{0} + \sum_{n=1}^{N} w_{nk}, \\
	\alpha_{k} & =  \alpha_{0} + \sum_{n=1}^{N} \sum_{l=k+1}^{K} w_{nl}.
\end{align*}

\bibliographystyle{IEEEtran}
\bibliography{references.bib}

\end{document}

%% file: packages.tex
\usepackage[T1]{fontenc}

\usepackage[nocompress]{cite}

\usepackage{dsfont}
\usepackage{amsfonts}
\usepackage{textcomp}

\usepackage{amsmath}
\usepackage{amssymb}
\usepackage{commath}
\usepackage{mathtools}

\usepackage{graphicx}
\usepackage{adjustbox}
\usepackage{caption}
\usepackage{svg}

\usepackage{pgfplots}
\usepackage{tikz}     
\usepackage{tikz-layers}

\pgfdeclarelayer{background}
\pgfdeclarelayer{middleground}
\pgfdeclarelayer{foreground}
\pgfsetlayers{background, main, middleground, foreground}

\pgfplotsset{compat=newest}
\usetikzlibrary{decorations.pathmorphing}
\usetikzlibrary{fit}
\usetikzlibrary{backgrounds}
\usetikzlibrary{pgfplots.groupplots}

\usepackage{url}
\usepackage{booktabs}
\usepackage[binary-units]{siunitx}
\usepackage{xcolor}
\usepackage{multirow}

\usepackage{hyperref}
\hypersetup{
	breaklinks=true,
	pdfpagelabels,
	colorlinks=true,
	linkcolor=blue,
	filecolor=magenta,      
	urlcolor=blue,
	citecolor=blue,
	linktoc=none,
}

\usepackage[acronym, shortcuts,]{glossaries}
\glsdisablehyper

%% file: macros.tex
\usepackage{bm}

\renewcommand{\vec}[1]{\bm{\mathbf{#1}}}
\newcommand{\mat}[1]{\bm{\mathbf{#1}}}

\newcommand{\hvec}[1]{\hat{\bm{\mathbf{#1}}}}

\DeclareMathOperator*{\argmin}{arg\,min}

\DeclareMathOperator{\const}{const}
\DeclareMathOperator{\kl}{KL}

\newcommand{\op}[1]{\operatorname{#1}}

\newcommand{\Dir}{\op{Dir}}
\newcommand{\Cat}{\op{Cat}}
\newcommand{\Beta}{\op{Beta}}

\newcommand{\N}{\op{N}}
\newcommand{\NW}{\op{NW}}

\newcommand{\MN}{\op{MN}}

\newcommand{\W}{\op{W}}

%% file: acronyms.tex
\newacronym{EM}{EM}{expectation-maximization}
\newacronym{RL}{RL}{reinforcement learning}
\newacronym{Hb-REPS}{Hb-REPS}{hybrid relative entropy policy search}
\newacronym{SDS}{SDS}{switching dynamical systems}
\newacronym{SSM}{SSM}{switching state-space models}
\newacronym{PWL}{PWL}{piecewise linear}
\newacronym{ReLU}{ReLU}{rectified linear units}
\newacronym{PWARX}{PWARX}{piecewise autoregressive exogenous systems}
\newacronym{MIQP}{MIQP}{mixed-integer quadratic program}
\newacronym{PGM}{PGM}{probabilistic graphical models}
\newacronym{HDBN}{HDBN}{hybrid dynamic Bayesian networks}
\newacronym{BNP}{BNP}{Bayesian nonparametrics}
\newacronym{MCMC}{MCMC}{Markov chain Monte Carlo}
\newacronym{SVI}{SVI}{stochastic variational inference}
\newacronym{HRL}{HRL}{hierarchical reinforcement learning}
\newacronym{SMDP}{SMDP}{semi-Markov decision processes}
\newacronym{MDP}{MDP}{Markov decision processes}
\newacronym{rARHMM}{rARHMM}{recurrent autoregressive hidden Markov model}
\newacronym{ARHMM}{ARHMM}{autoregressive hidden Markov model}
\newacronym{HMM}{HMM}{hidden Markov model}
\newacronym{HSMM}{HSMM}{semi-hidden Markov model}
\newacronym{SLDS}{SLDS}{switching linear dynamical systems}
\newacronym{rSLDS}{rSLDS}{recurrent switching linear dynamical systems}
\newacronym{FHMM}{FHMM}{factorial hidden Markov model}
\newacronym{REPS}{REPS}{relative entropy policy search}
\newacronym{KL}{KL}{Kullback-Leibler divergence}
\newacronym{MAP}{MAP}{maximum a posteriori}
\newacronym{NW}{NW}{normal-Wishart}
\newacronym{MN}{MN}{matrix-normal}
\newacronym{MNW}{MNW}{matrix-normal-Wishart}
\newacronym{Dir}{Dir}{Dirichlet}
\newacronym{Cat}{Cat}{categorical}
\newacronym{E-step}{E-step}{expectation step}
\newacronym{M-step}{M-step}{maximization step}
\newacronym{EB-step}{EB-step}{empirical Bayes step}
\newacronym{FNN}{FNN}{feed-forward neural net}
\newacronym{GP}{GP}{Gaussian process}
\newacronym{LSTM}{LSTM}{long-short-term memory network}
\newacronym{RNN}{RNN}{recurrent neural network}
\newacronym{NMSE}{NMSE}{normalized mean squared error}
\newacronym{SAC}{SAC}{soft actor-critic}
\newacronym{RFF}{RFF}{random Fourier feature}
\newacronym{GPR}{GPR}{Gaussian process regression}
\newacronym{SGPR}{SGPR}{sparse Gaussian process regression}
\newacronym{ANN}{ANN}{artificial neural network}
\newacronym{LR}{LR}{local regression}
\newacronym{ILR}{ILR}{infinite local regression}
\newacronym{HILR}{HILR}{hierarchical infinite local regression}
\newacronym{CMB}{CMB}{cosmic microwave background}
\newacronym{BNN}{BNN}{Bayesian neural network}
\newacronym{RFWR}{RFWR}{receptive field weighted regression}
\newacronym{LWPR}{LWPR}{locally weighted projection regression}
\newacronym{LGR}{LGR}{local Gaussian regression}
\newacronym{MoE}{MoE}{mixture of experts}
\newacronym{VBEM}{VBEM}{variational Bayes expectation-maximization}
\newacronym{VB}{VB}{variational Bayes}
\newacronym{GMM}{GMM}{Gaussian mixture model}
\newacronym{DP}{DP}{Dirichlet process}
\newacronym{VI}{VI}{variational inference}
\newacronym{ELBO}{ELBO}{evidence lower bound}
\newacronym{MSE}{MSE}{mean squared error}
\newacronym{PoE}{PoE}{product of experts}
\newacronym{gPoE}{gPoE}{generalized product of experts}
\newacronym{rBCM}{rBCM}{robust Bayesian committee machine}

%% file: figures/triangle.tex
\includegraphics{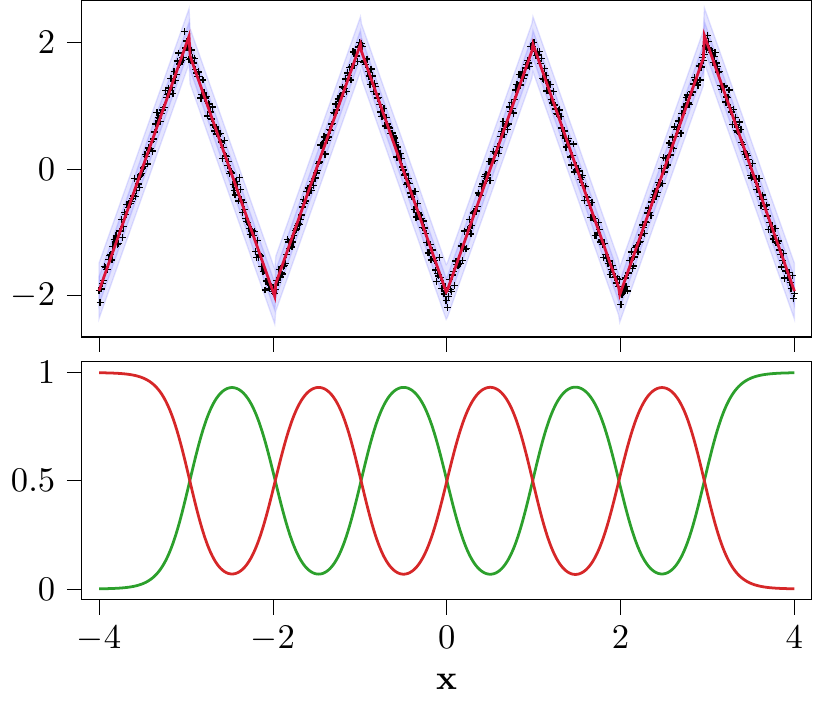}

%% file: figures/step.tex
\includegraphics{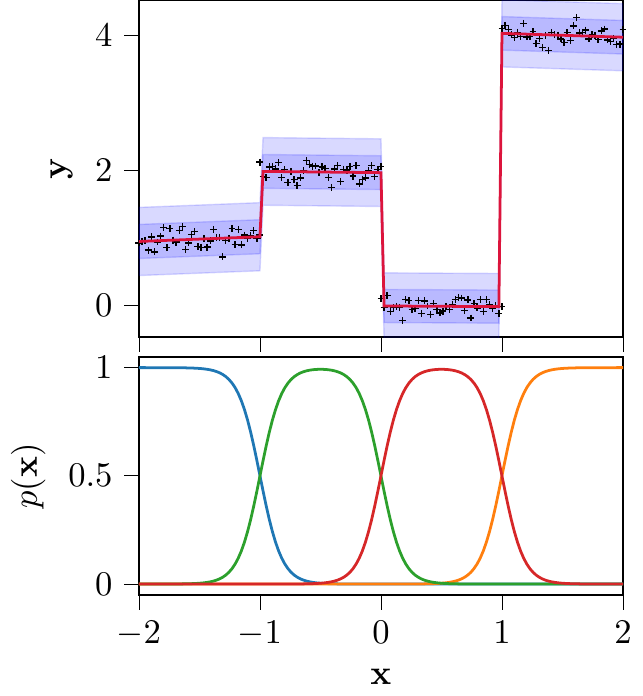}

%% file: figures/step_poly_features.tex
\includegraphics{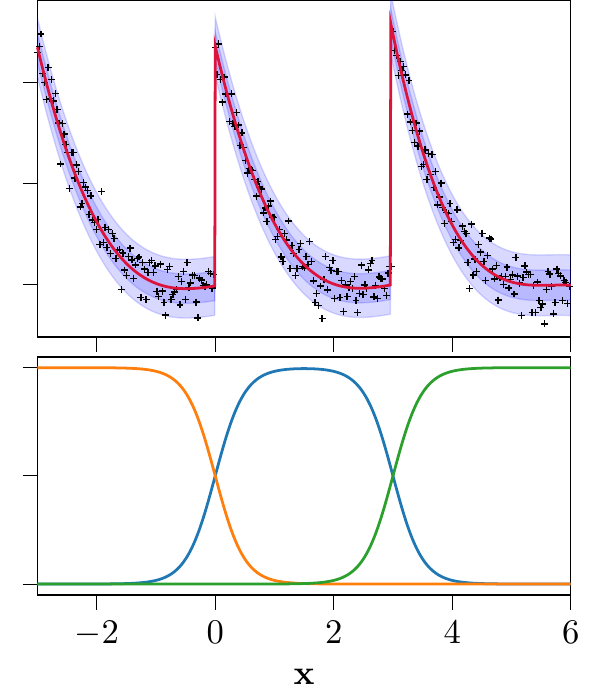}

%% file: figures/inverse_experts.tex
\includegraphics{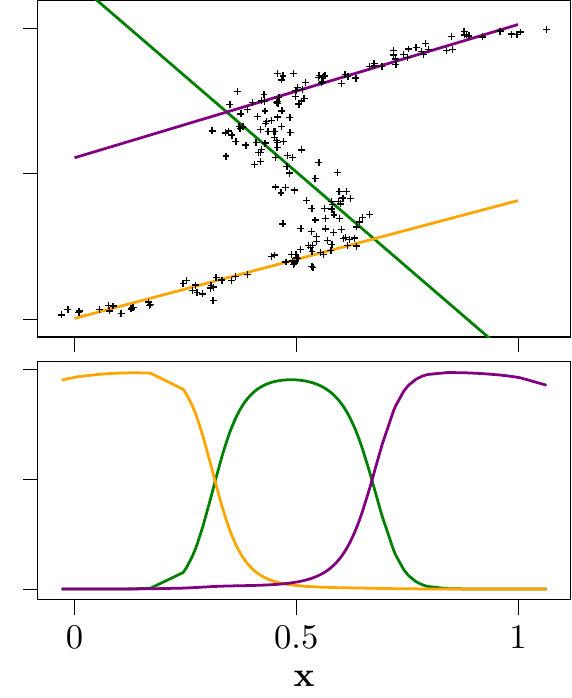}

%% file: figures/chirp_0.tex
\includegraphics{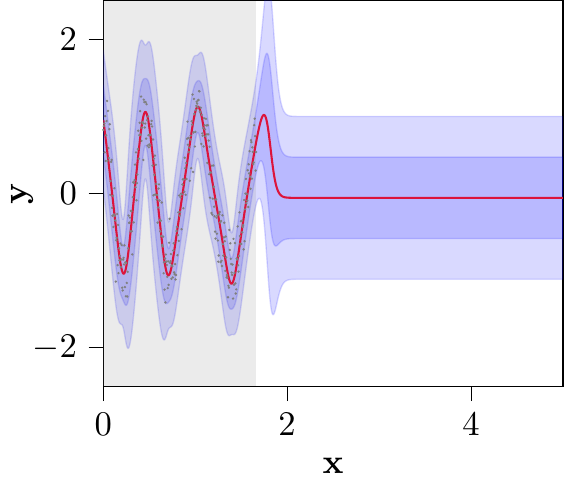}

%% file: figures/chirp_1.tex
\includegraphics{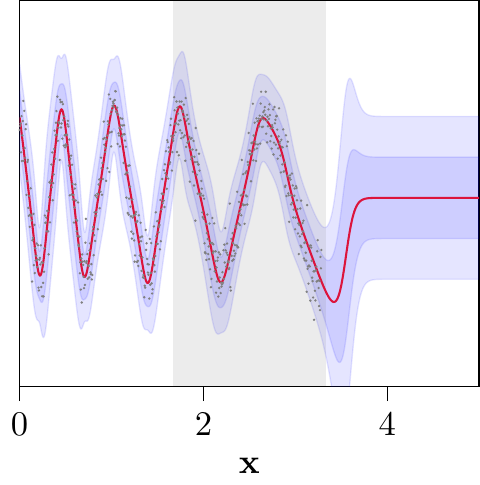}

%% file: figures/chirp_2.tex
\includegraphics{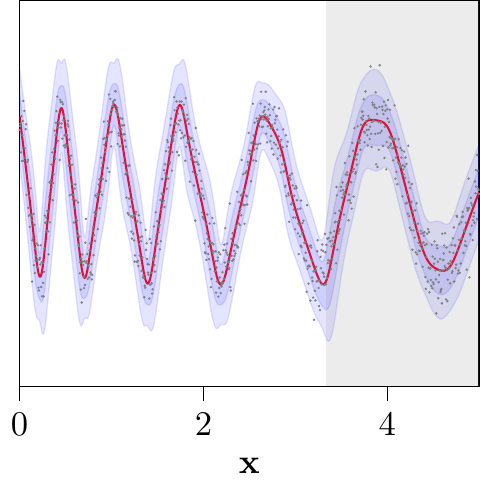}

%% file: figures/eight_seq_smse.tex
\includegraphics{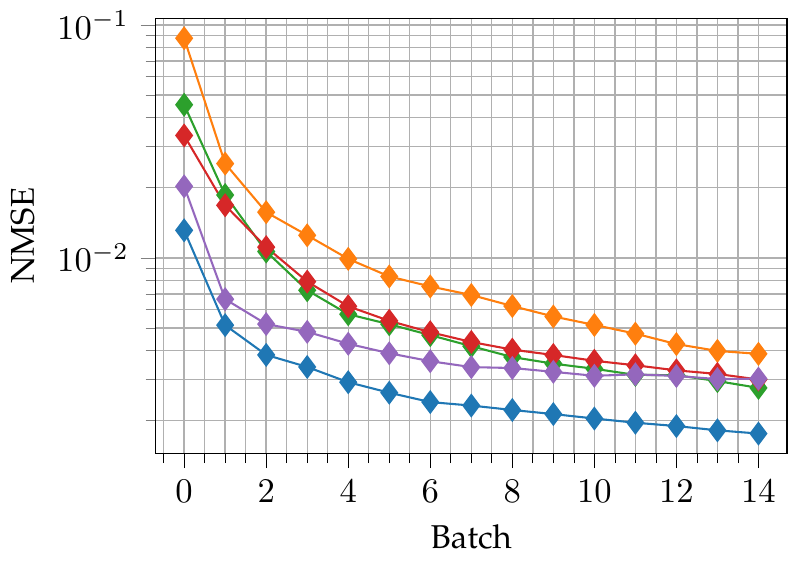}

%% file: figures/eight_0275.tex
\includegraphics{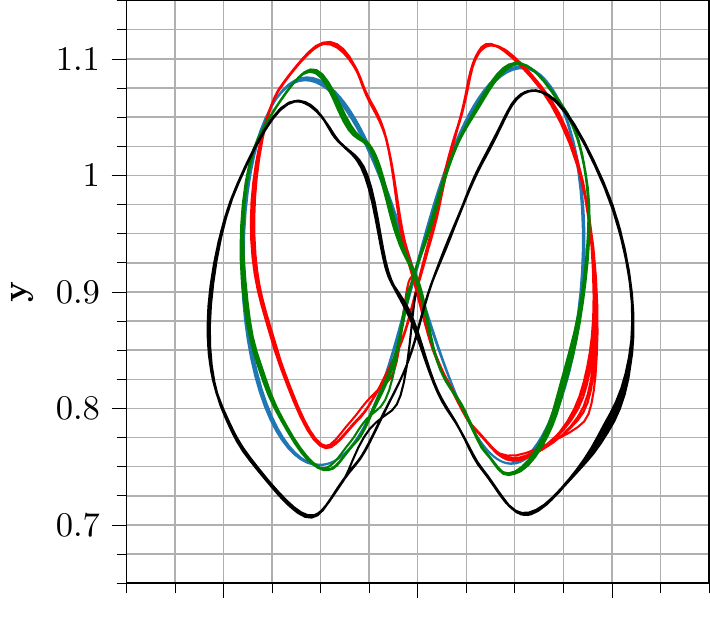}

%% file: figures/eight_0425.tex
\includegraphics{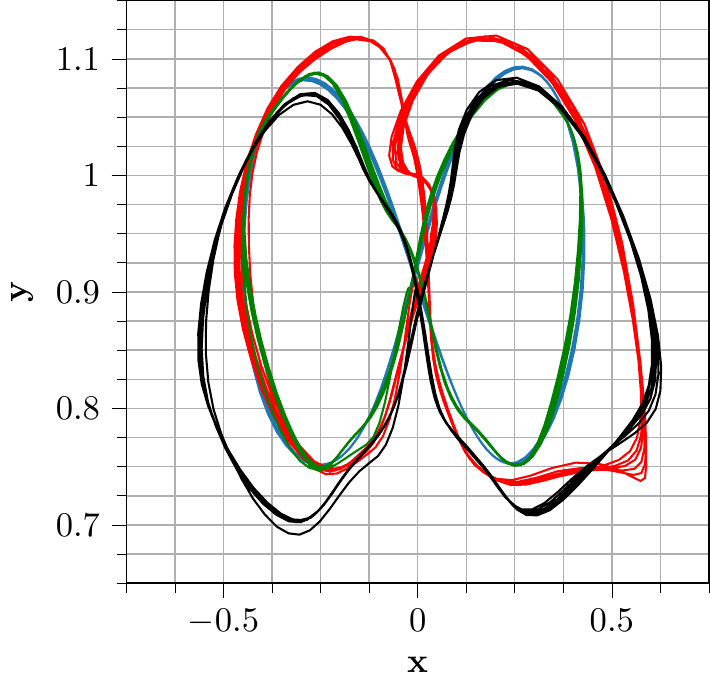}